\documentclass{article}

\usepackage{arxiv}

\usepackage[utf8]{inputenc} 
\usepackage[T1]{fontenc}    

\usepackage{hyperref}       
\usepackage{url}            
\usepackage{booktabs}       
\usepackage{amsfonts}       
\usepackage{nicefrac}       
\usepackage{microtype}      
\usepackage{lipsum}		

\usepackage{amssymb}
\usepackage{todonotes}
\usepackage{float}
\usepackage{tabularx}
\usepackage{booktabs}
\usepackage{multirow}
\usepackage{makecell}
\usepackage{utfsym} 
\usepackage{url}
\usepackage{amsmath}

\usepackage{threeparttable}
\usepackage{algorithm2e}

\usepackage[table]{xcolor} 
\usepackage{adjustbox}
\usepackage{caption,subcaption}   

\usepackage{natbib}
\usepackage{doi}
\usepackage{xspace}
 \usepackage{soul}
\usepackage[normalem]{ulem} \usepackage{xcolor}

\newcommand{\highlightyellowrevnew}[1]{{#1}}

 \newcommand{\hlsecondround}[1]{{#1}}

\newcommand{\cmark}{\usym{1F5F8}}
\newcommand{\xmark}{\usym{2717}}

\title{Spot-Wise Smart Parking: An Edge-Enabled Architecture with YOLOv11 and Digital Twin Integration}

\author{
    Gustavo P. C. P. da Luz \\
    Institute of Computing\\
    University of Campinas (UNICAMP)\\
    Av. Albert Einstein, 1251, \\
    Campinas, 13083-852, SP, Brazil \\
    \texttt{ra271582@students.ic.unicamp.br} \\
    \And
    Alvaro M. Aspilcueta Narvaez \\
    Institute of Computing\\
    University of Campinas (UNICAMP)\\
    Av. Albert Einstein, 1251, \\
    Campinas, 13083-852, SP, Brazil \\
    \texttt{a272497@dac.unicamp.br} \\
    \And
    Tiago Godoi Bannwart \\
    Institute of Computing\\
    University of Campinas (UNICAMP)\\
    Av. Albert Einstein, 1251, \\
    Campinas, 13083-852, SP, Brazil \\
    \texttt{t215386@dac.unicamp.br} \\
    \And
    Gabriel Massuyoshi Sato \\
    Institute of Computing\\
    University of Campinas (UNICAMP)\\
    Av. Albert Einstein, 1251, \\
    Campinas, 13083-852, SP, Brazil \\
    \texttt{ra172278@students.ic.unicamp.br} \\
    \And
    Luis Fernando Gomez Gonzalez\thanks{Corresponding Author}\\
    Institute of Computing\\
    University of Campinas (UNICAMP)\\
    Av. Albert Einstein, 1251, \\
    Campinas, 13083-852, SP, Brazil \\
    \texttt{gonzalez@unicamp.br} \\
    \And
    Juliana Freitag Borin\\
    Institute of Computing\\
    University of Campinas (UNICAMP)\\
    Av. Albert Einstein, 1251, \\
    Campinas, 13083-852, SP, Brazil \\
    \texttt{juliana@ic.unicamp.br} \\
}

\renewcommand{\shorttitle}{Spot-Wise Smart Parking Edge-Enabled with YOLOv11 Towards a Digital Twin}

\hypersetup{
pdftitle={Spot-Wise Smart Parking: An Edge-Enabled Architecture with YOLOv11 and Digital Twin Integration},
pdfsubject={q-bio.NC, q-bio.QM},
pdfauthor={Gustavo P.C.P da Luz},
pdfkeywords={First keyword, Second keyword, More},
}

\begin{document}
\maketitle

\begin{abstract}
    Smart parking systems help reduce congestion and minimize \hlsecondround{users'} search time, thereby contributing to smart city adoption and enhancing urban mobility. In previous works, we presented a system developed on a university campus to monitor parking availability by estimating the number of free spaces from vehicle counts within a region of interest. Although this approach achieved good accuracy, it restricted the system’s ability to provide spot-level insights and support more advanced applications. To overcome this limitation, we extend the system with a spot-wise monitoring strategy \highlightyellowrevnew{based on a distance-aware matching method with spatial tolerance}, enhanced through an Adaptive Bounding Box Partitioning method for challenging spaces. The proposed approach achieves a balanced accuracy of \highlightyellowrevnew{98.80}\% while maintaining an inference time of 8 seconds on a resource-constrained edge device, \highlightyellowrevnew{enhancing the capabilities of YOLOv11m, a} model that has a size of 40.5 MB. In addition, two new components were introduced: (i) a Digital \hlsecondround{Shadow} that visually represents parking lot entities \hlsecondround{as a base to evolve to a full Digital Twin}, and (ii) an application support server based on a repurposed TV box. The latter not only enables scalable communication among cloud services, the parking totem, and a bot that provides detailed spot occupancy statistics, but also promotes hardware reuse as a step towards greater sustainability.
 
\end{abstract}

\keywords{Smart Parking \and Internet of Things \and Deep Learning \and Digital Twin \and TV Box.}
\section{Introduction}
\label{sec:introducao}

\highlightyellowrevnew{Efficient parking management is a key challenge in modern urban environments, where drivers spend a significant portion of their travel time searching for available spaces, accounting for up to 30\% of urban traffic} \cite{shoup2021pricing} \highlightyellowrevnew{and contributing substantially to fuel waste and carbon emissions} \cite{rodriguez2024analysis,paidi2022co2}. 

\highlightyellowrevnew{Improving parking utilization directly supports the United Nations Sustainable Development Goal (SDG) 11, which promotes the creation of inclusive, safe, resilient, and sustainable cities} \cite{nations2015transforming}. \highlightyellowrevnew{Furthermore, smart parking systems have the potential to improve the user experience and generate valuable data for urban planning and the development of more integrated and sustainable environments}~\cite{de2024smart}.

Aligned with these goals, Unicamp has launched the Smart Campus initiative\footnote{\href{https://smartcampus.prefeitura.unicamp.br/}{Smart Campus Unicamp}}, aiming to transform the campus into a more intelligent, efficient, and sustainable environment by implementing Internet of Things (IoT) technologies. These technologies seek to optimize decision-making processes and improve both productivity and quality of life in the university environment. One of the proposals of Smart Campus is the implementation of smart parking systems. Given the high demand for parking spaces at the Unicamp campuses, the system aims to reduce the time drivers spend searching for an available spot \highlightyellowrevnew{and contribute to sustainability by lowering greenhouse gas emissions and fuel consumption.}

Traditional smart parking systems often rely on infrared, ultrasonic, or magnetic sensors, \highlightyellowrevnew{which are costly and unsuitable for outdoor areas. Vision-based solutions, in contrast, offer a more cost-effective and easier to install and maintain alternative by using cameras to monitor multiple spaces simultaneously. However, the deployment of deep learning models on edge devices such as the Raspberry Pi remains challenging due to limited computational resources. Edge intelligence addresses this issue by running optimized AI models locally, reducing network latency, minimizing data transmission, and preserving user privacy,} as sensitive image data never leaves the local device. In practice, pre-trained models on large datasets can be adapted for edge deployment, achieving intelligent monitoring while maintaining efficiency and privacy.

In this work, we present a near real-time monitoring system for the number of available parking spaces in the Institute of Computing building (building IC-2) at Unicamp, adopting what we call a spot-wise approach. Unlike previous versions of the system, which only reported the aggregate number of cars and available spaces \cite{da202510}, the new approach identifies the status of each individual parking spot (occupied or vacant). This extension not only increases the granularity of the information provided but also opens up new opportunities for system integration, user services, and data-driven management of parking resources. Building on this spot-wise approach, this paper makes the following key contributions:

\begin{itemize}
\item system architecture: description of the new system layers, including the deployment of an application support server hosted on a repurposed TV Box and details of the totem hardware installation.
\item detection methods: implementation of two complementary methods to enable and improve spot-level occupancy detection.
\item \hlsecondround{Digital Twin integration: implementation of a foundational Digital Shadow component aligned with Smart City standards, establishing the initial step toward a full Smart Campus Parking Digital Twin.}

\item performance evaluation: a statistical analysis of daily and monthly spot occupancy, providing insights into usage patterns and system effectiveness.
\end{itemize}

\hlsecondround{Given the varied and sometimes inconsistent terminology in the literature, it is helpful to first distinguish a Digital Shadow (DS) from a Digital Twin (DT). A DS is typically defined as a virtual representation that receives automated, one-way data updates from its physical counterpart and does not exert control over the physical asset. By contrast, a DT denotes a system with integrated, automated two-way communication: it not only mirrors the physical object but can exchange information in real time and potentially influence, optimize, or control the physical system. Under this distinction, the presence of bidirectional, often real-time data exchange is the defining characteristic that elevates a DS into a DT} \cite{bergs2021concept, sepasgozar2021differentiating, brarda2025digital, silva2025not}.

\hlsecondround{In this work we adopt the above definition: our implementation provides a structured, automated digital representation of parking monitoring system without real-time actuated control, thereby establishing a foundation for future evolution toward higher DT maturity.}

The remainder of the paper is organized as follows: Section \ref{sec:relacionados} reviews related work and the previous versions of the system, including architectures, functionalities, and limitations. Section \ref{sec:proposta} details the proposed system and its improvements. Section \ref{sec:avaliacao} presents the results obtained. Finally, Section \ref{sec:conclusao} concludes the paper and suggests directions for future work.

\section{Related Work and our Previous System Versions}
\label{sec:relacionados}

This section reviews related work in smart parking systems and summarizes previous versions of our own system, highlighting how the current spot-wise, edge-deployed approach advances and differs from these prior efforts.

\subsection{Related Work}
Research in deep learning-based smart parking has evolved along several key aspects. Amato et al.\cite{amato2017deep} introduced one of the first systems in this field, releasing the CNRPark-\highlightyellowrevnew{EXT} dataset and deploying their solution on a Raspberry Pi 2 edge device. Their approach \highlightyellowrevnew{trained models inspired by LeNet (AlexNet, mAlexNet) as well as Support Vector Machines (SVMs). To monitor individual parking spots within a defined region of interest (ROI), they used manually built masks to crop each input image into frames corresponding to each one of the parking spots before they are fed into the classifier model. Similarly,} Nyambal and Klein\cite{nyambal2017automated} \highlightyellowrevnew{also trained LeNet and AlexNet models for spot-level monitoring at a dataset of two parking lots of the
University of the Witwatersrand based on images of each parking spot frame. To crop each spot they stored the cartesian coordinates of each spot in an Extensible Markup Language (XML) file generated by a labeling tool. In their case, they adopted a cloud-based architecture for the deploy and did not use an edge device
for the inference.}

Subsequent works explored more modern deep learning models but were limited to evaluations on public or benchmark datasets, without considering edge devices or real-world deployments. For instance, Abbas et al.\cite{abbas2023revolutionizing} applied YOLOv3 and AlexNet models finetuned on the PKLot dataset for parking space monitoring, evaluating performance at the frame level for each parking spot. \highlightyellowrevnew{The dataset} \cite{de2015pklot}\highlightyellowrevnew{ includes cropped sub-images for each parking spot, extracted according to coordinates specified in an XML file generated by a labeling tool.}
Although \highlightyellowrevnew{all the mentioned studies adopted a spot-wise approach, training models on individual parking spot frames} \highlightyellowrevnew{is more computationally expensive than performing a single inference on the entire image containing all parking spots.} \hlsecondround{Another common limitation is the absence of pre-trained models, advanced system management, and the use of Digital Shadows for data visualization.} Pre-trained models are particularly advantageous because they leverage knowledge learned from large-scale datasets, reducing the amount of task-specific data required, improving generalization in real-world scenarios, and accelerating application development, \highlightyellowrevnew{since training is not required}. By comparison, Shrivastav et al.\cite{shrivastav2023parking} employed a pre-trained YOLOv5 model and deployed their system on a local municipality server in the city of Modena, using live video feeds from real parking lots. In addition to YOLO-based detection, DeepSORT was integrated for vehicle tracking, and a 3D Digital \hlsecondround{Shadow (according to our used definition)} was developed to monitor and visualize parking spaces through a spot-wise approach. \highlightyellowrevnew{The method used was to treat all detected vehicles as potential parking locations. Their image coordinates and unique identifiers are stored to form a database of parking spots for each camera. During inference, the corresponding ROI is retrieved from this database, cropped from the original frame, and passed to a classifier to predict the occupancy status of each parking space.} While this represents a modern solution leveraging 5G high-bandwidth infrastructure, inference was still performed on a cloud server rather than directly on an edge device. Another limitation is that if the image contains cars parking outside the parking lot they will still be counted as vehicles.

More recently, Luong et al.\cite{luong2025ai} employed a finetuned YOLOv11x model to detect available and occupied parking spots in a university parking lot. The system includes a web application displaying a map of the parking lot with spots marked as empty or occupied, \hlsecondround{which we considered as a Digital Shadow}. Despite the real-world deployment, all inference was performed on cloud servers, and the system did not leverage edge computing for on-site processing. 
\highlightyellowrevnew{The spot-wise method used operates by generating a vehicle mask for each frame and determining occupancy by checking for overlap with predefined parking coordinates. A limitation of this approach is that it relies on exact spatial alignment and no distance threshold is considered when the detected vehicle position deviates slightly from the defined parking spot.}

\hlsecondround{Although earlier approaches offer valuable information to end users, the absence of interoperability standards results in scalability issues and restricted integration with wider smart city infrastructures. Our work provides a Digital Shadow of the parking system and addresses these limitations through the use of NGSI-LD-compliant Smart Data Models within the FIWARE framework, ensuring interoperability with other smart city solutions.}

As mentioned earlier, most of the reviewed works perform computational tasks in the cloud. In contrast, our approach performs inference on a Raspberry Pi connected to the camera capturing the images, while also leveraging a repurposed TV Box as a support server for the smart parking system. To the best of our knowledge, no previous proposal \highlightyellowrevnew{from other research groups} has explored the use of repurposed devices in this role, as support servers are traditionally implemented with cloud services or personal computers. TV Boxes, often apprehended by regulatory agencies, have been successfully repurposed to provide innovative solutions that also contribute to the circular economy. Prior studies have used TV Boxes for people counting \cite{courb,sato2025people}, fall detection \cite{shu2021eight}, leaf disease prediction \cite{moreira2022agrolens}, educational applications \cite{campos2025projeto,pinhao2025explorando}, and resilience testing under stress conditions \cite{da2024repurposing}. However, in these works the TV Box served either as the inference device or as a digital learning platform, whereas in our case it is employed as an application support server for a real-world smart parking deployment. The detailed implementation of the server is described in \cite{bannwart2025tv}.

Most smart parking solutions rely on state-of-the-art models and advanced applications; however, they are not designed for compatibility with edge devices, which could enhance such systems by improving privacy and reducing latency.

\subsection{Our Previous System Versions}

\begin{table*}[htpb]
\centering
\begin{threeparttable}
\begin{tabular}{@{}p{3.9cm} c c c c c  c@{}}
\toprule
\textbf{Work} & \makecell{\textbf{Edge} \\ \textbf{Inf.}}  & \textbf{Best Model} & \makecell{\textbf{Pre} \\ \textbf{Trained}}  & \textbf{Spots} & \makecell{\textbf{Digital} \\ \hlsecondround{\textbf{Shadow}}} & \textbf{Deploy} \\
\midrule
Amato et al.\cite{amato2017deep} & \cmark & mAlexNet & \xmark & \cmark & \xmark & \cmark \\
Nyambal and Klein\cite{nyambal2017automated} & \xmark & LeNet & \xmark &  \cmark & \xmark & \cmark \\
Abbas et al.\cite{abbas2023revolutionizing} & \xmark & YOLOv3 & \xmark &\cmark & \xmark & \xmark \\
Shrivastav et al.\cite{shrivastav2023parking} & \xmark & YOLOv5 & \cmark  & \cmark & \cmark & \cmark \\
Luong et al.\cite{luong2025ai}& \xmark & YOLOv11x & \xmark  & \cmark & \cmark & \cmark \\
\midrule
\multicolumn{7}{c}{\textbf{Our Works}} \\
\midrule
2015 phase & \xmark & LeNet & \xmark & \cmark & \xmark & \xmark \\
2019 phase & \cmark & EfficientDet D2 Lite & \cmark& \xmark & \xmark & \cmark \\
Baggio et al.\cite{SmartParkingUnicamp} & \cmark & YOLOv3 & \cmark &  \xmark & \xmark & \xmark \\
da Luz et al.\cite{da2024smart} & \cmark & YOLOv8--v11 & \cmark & \xmark & \xmark & \xmark \\
da Luz et al.\cite{da202510} & \cmark & YOLOv11m TFLite & \cmark & \xmark & \xmark & \cmark\tnote{1} \\
\textbf{This Work (2026)} & \cmark & YOLOv11m TFLite & \cmark &  \cmark & \cmark & \cmark\tnote{2} \\
\bottomrule
\end{tabular}
\begin{tablenotes}
      \footnotesize
      \item[1] Deploy at Raspberry Pi.
      \item[2] Deploy at Raspberry Pi and Totem.
\end{tablenotes}
\end{threeparttable}
\caption[Comparison of deep learning-based Smart Parking Approaches]{Comparison of deep learning-based smart parking approaches from the literature and our research group, considering edge-based inference, model type, use of pre-trained models, adoption of a spot-wise approach, presence of a Digital \hlsecondround{Shadow}, and deployment status.}
\label{tab:related-works}
\end{table*}

Our research on smart parking systems began around 2015, inspired by advances in deep learning. Initially, we experimented with the GoogleLeNet architecture \cite{szegedy2015going}, employing a teacher-student distillation approach where a large pre-trained Xception network \cite{chollet2017xception} generated annotations to train a smaller student model. At this stage, a spot-wise approach was adopted, with the model trained per parking spot, using bounding boxes generated from the intersection between the segmentation map and predefined parking spots, with annotations assigned as cars when the overlap exceeded 50\%. A more detailed discussion of this method was presented at an international conference on real-world machine learning applications\footnote{\href{https://www.youtube.com/watch?v=vRXgc0Bvbx8}{PAPIs.io
 LATAM 2018 – Full Conference Day 1}}. This foundational work, while resulting in a proof-of-concept, was not deployed on edge hardware and remained a research prototype.

The transition from research to an operational system occurred in 2019 with our first successful deployment. With the challenge of running deep learning models on resource-constrained devices, we used the Single Shot Detector (SSD) pre-trained EfficientDet-D2 Lite \cite{tan2020efficientdet} with a deployment on a Raspberry Pi 3B+, which is still used in the parking lot. At this point, the spot-wise approach was discontinued, as a pre-processing ROI method was used to select the parking lot area where cars are parking. This method was based on a predefined reference mask used to keep only the parking lot region at the image prior to model inference. The same method was used in the evaluation phase that took place from 2020-2024, building on the deployed foundation. At this 2020 phase, we conducted comparative studies of various model families, including an analysis of a pre-trained YOLOv3 model \cite{redmon2018yolov3, SmartParkingUnicamp} which demonstrated promising results for our application context. Although not deployed in a real-world setting, the solution was tested on Raspberry Pi 3 and Raspberry Pi 4 devices for running the inference tasks.

The evaluating phase from 2020-2024 resulted on a benchmarking study in 2024 \cite{da2024smart} that evaluated the latest pre trained YOLO generations models (v8 to v11) across multiple edge and cloud devices. This research provided the basis for model selection and incurred on our subsequent deployment of a YOLOv11m-TFLite model on the Raspberry Pi edge device \cite{da202510}. We also evaluated the selected model comparing it to 11 different models, comprehending EfficientDet-D2 Lite, Mask R-CNN, YOLOv3, and YOLOv8-v11. These works were based on a new post-processing selection method that enables ROI detection after the inference, keeping relevant context information to the deep learning models. Despite the ROI usage, up to this point the system performed general vehicle detection without monitoring individual parking spaces.

The present work, also deployed in 2025, represents the extension of the work \cite{da202510}, integrating the optimized YOLOv11m-TFLite model into a cyber-physical system that advances the other approaches. Our system now performs accurate per-space monitoring of all parking spots within a defined region of interest at a Raspberry Pi as edge device, creates a real-time interoperable standardized \hlsecondround{Digital Shadow} for system visualization and management, and incorporates a dedicated application support server hosted on a repurposed TV Box. \highlightyellowrevnew{Unlike prior spot-wise methods, which determine occupancy by cropping the full image into singular parking spots or by checking for exact overlap between vehicle masks and predefined parking coordinates, our approach introduces a distance-based matching method with spatial tolerance that can be used with pre-trained models and inference at the edge. Specifically, for each detection, we compute the Euclidean distance between the detection center and every parking spot center, assigning the nearest spot within a threshold distance. Combined with a pixel-wise ROI selection, this method enhances robustness against small positional deviations and improves the reliability of spot assignment. We further evaluate the impact of incorporating this spot-wise strategy on balanced accuracy and inference time compared to previously considered models.}

Table \ref{tab:related-works} summarizes how each of the reviewed works, as well as our previous systems, operate in terms of edge-device inference, inference model, use of pre-trained models, adoption of a spot-wise approach, \hlsecondround{Digital Shadow towards a Digital Twin} integration, and deployment status.


\section{Proposed System}
\label{sec:proposta}

The current system can be described using the standard layered architecture commonly employed in IoT systems.

\begin{figure}[htpb]
  \centering
  \includegraphics[width=0.39\textwidth]{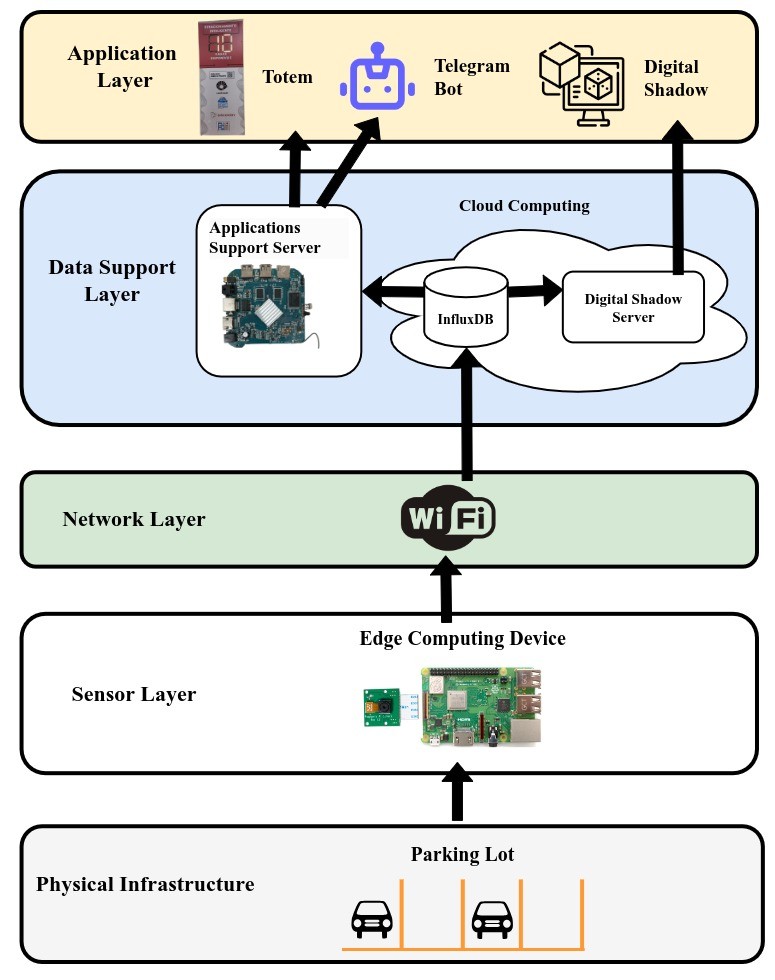} 
  \caption{IoT layers of the proposed system. \highlightyellowrevnew{Component image sources:} \href{https://commons.wikimedia.org/wiki/File:Raspberry_Pi_3_B\%2B_\%2839906369025\%29.png}{Raspberry Pi 3B+} and
  \href{https://gopigo.io/raspberry-pi-camera/}{Raspberry Pi Camera Module}.}
  \label{fig:layers}
\end{figure}

Figure \ref{fig:layers} illustrates the system model, describing the complete data flow from the capture of a new image to the display of information on the totem LED panel. In the sensor layer, a Raspberry Pi is connected to its standard camera module, which captures photos of the parking lot, processes them, and transmits the parking status to InfluxDB over Wi-Fi.

In the data support layer, the InfluxDB time-series database and the \hlsecondround{digital shadow} server are hosted in the cloud. In addition to these two components, an application support server was included to provide parsed data from InfluxDB to the totem and to host the monitoring bot. 

Finally, applications such as the totem, the Telegram bot, and the \hlsecondround{digital shadow} provide the parking status to end users and stakeholders. The next sections of this paper are organized to follow the main components of Figure \ref{fig:layers}, from the bottom to the top, describing in more details the architecture of its components.

\subsection{Edge Device}

\begin{figure}[htpb]
  \centering
  \includegraphics[width=0.52\textwidth]{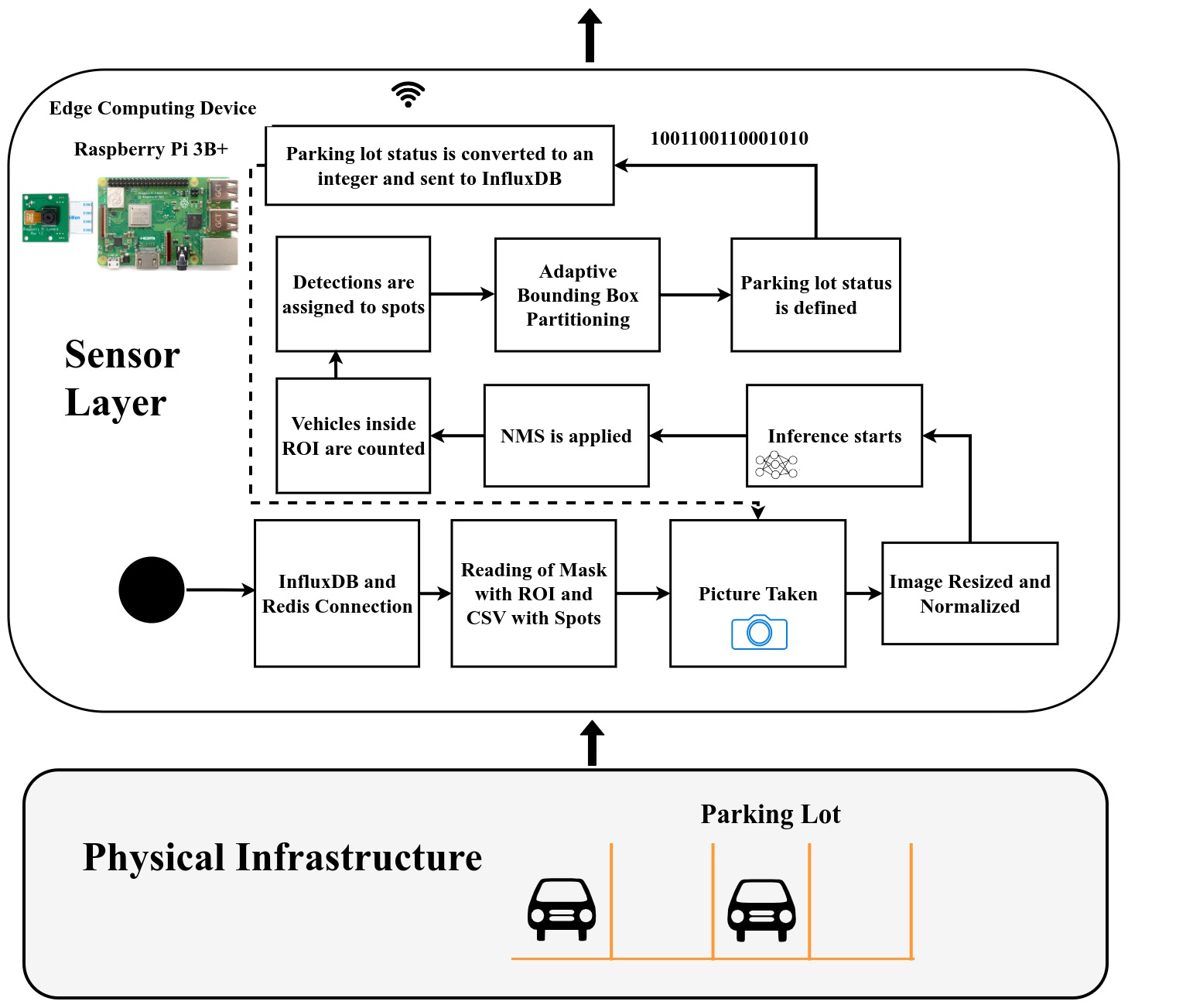} 
  \caption{Edge Computing Device Flow. \highlightyellowrevnew{The circle indicates the start of the script and the dashed arrow indicates the loop.}}
  \label{fig:flowpi}
\end{figure}

The first component is the edge device, responsible for collecting data from the physical parking lot infrastructure. Figure \ref{fig:flowpi} illustrates the steps executed on the edge device, implemented on a Raspberry Pi. 

The process begins by connecting the local Redis database to the cloud-based InfluxDB database. Redis temporarily stores data in a queue before transmitting it via Wi-Fi.

\begin{figure}[htpb]
  \centering
  \includegraphics[width=0.49\textwidth]{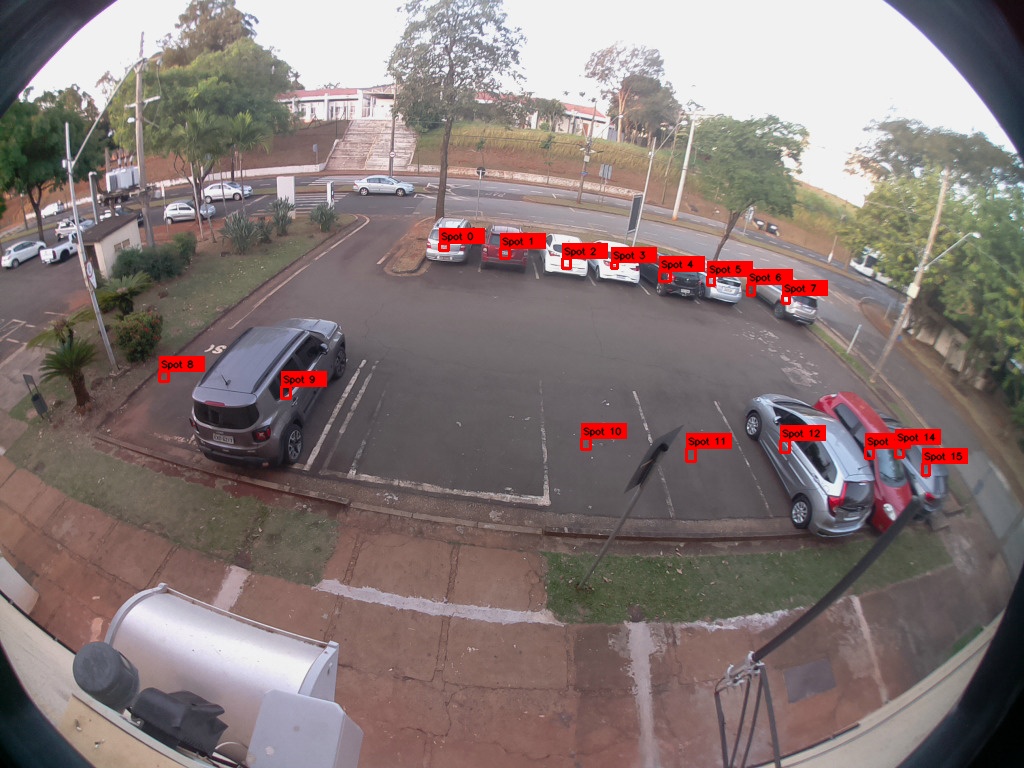} 
  \caption{Labeling of the center coordinates of each spot.}
  \label{fig:annot}
\end{figure}

Next, two input files are loaded from persistent storage into memory.  The first is a custom mask that enables pixel-wise selection of regions of interest (ROIs) used to count vehicles within designated parking areas during post-processing. The second is a comma-separated values (CSV) file that specifies the location of each parking spot. Each row corresponds to a labeled image and contains a serialized list of YOLO-format bounding boxes predicted for that reference image. Although this format can be used as an annotation tool for multiple images, in our case it was applied to a single image to obtain the annotations for all parking spots. The bounding boxes, which store the individual coordinates of each parking spot, are illustrated in Figure \ref{fig:annot}. The squares represent the center of each spot and were generated using a Python script with a graphical user interface (GUI).

After the input files are loaded, the system enters a continuous loop in which new images of the parking lot are captured and preprocessed for inference. The preprocessing includes normalization and resizing to match the input dimensions required by each machine learning model. Specifically, YOLOv3 expects an input of 416$\times$416 pixels, EfficientDet-D2 Lite uses 448$\times$448, Mask R-CNN uses 1024$\times$1024, and the YOLOv8–v11 models use 640$\times$640. The choice of input resolution has a direct impact on both model size and detection performance.

Once inference is completed, Non-Maximum Suppression (NMS) is applied as a post-processing step to retain only the most confident detections, effectively removing duplicate bounding boxes for the same object. In addition to NMS, three complementary post-processing steps are applied: (i) assigning detected vehicles to their corresponding parking spots within the ROI; (ii) applying an Adaptive Bounding Box Partitioning method to refine spatial associations; and (iii) representing parking occupancy in a compact data format for efficient transmission and storage.


\subsubsection{Assigning Vehicles Detections to Spots within ROI selection}

After applying NMS to filter the relevant predictions, the first method aims to assign each detected vehicle to its corresponding parking spot.

\begin{algorithm}[H]
\caption{Assign detections to parking spots}
\label{alg:assign}
\DontPrintSemicolon
\KwIn{Detections $D$, Spots $S$, image width $W$, image height $H$, mask $M$, max distance $\delta$}
\KwOut{Updated spots $S$ (with occupancy flag), assignment list $A$}

$A \gets \varnothing$\;
\ForEach{spot $s \in S$}{
    $s.\text{occupied} \gets \text{False}$\;
}
\ForEach{detection $d_i = (x_c,y_c,w,h) \in D$}{
    $nearestSpot \gets \text{None}$, \quad $minDist \gets \infty$\;
    \ForEach{spot $s_j \in S$}{
        $dist \gets \sqrt{(x_c - s_j.x_c)^2 + (y_c - s_j.y_c)^2}$\;
        \If{$(dist < minDist) \wedge (dist < \delta)$}{
            $minDist \gets dist$\;
            $nearestSpot \gets j$\;
        }
    }
    \If{$nearestSpot \neq \text{None}$}{
        \If{$ROIMethod(x_c,y_c,M) = 1$}{
            $S[nearestSpot].\text{occupied} \gets \text{True}$\;
            Append $(i, nearestSpot)$ to $A$\;
        }
        \Else{
            \tcp{Detection outside mask --- skipped}
        }
    }
}
\Return{$(S, A)$}\;
\end{algorithm}

Algorithm 1 describes the method used to assign YOLO detections to each corresponding annotated parking spot. We start with all spot occupancy flags empty, and for each detection, we compute the Euclidean distance between the detection center and every spot center. The nearest spot within a threshold distance $\delta$ is selected. If the detection also lies inside the ROI mask, the spot is marked as occupied and the pair (detection index, spot index) is recorded. We considered a $\delta$ of 0.1 after visual inspection of cars parked close and distant to each parking spot

\begin{figure}[htpb]
  \centering
  \includegraphics[width=0.49\textwidth]{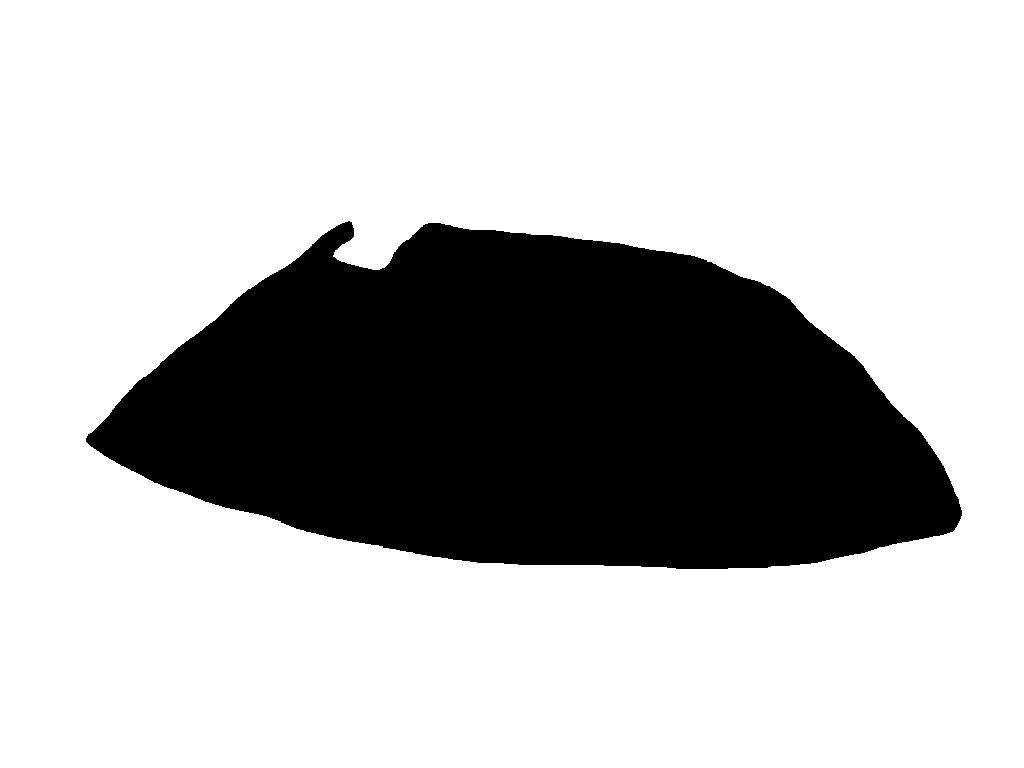} 
  \caption{ROI reference mask used to select the parking lot area.}
  \label{fig:roimask}
\end{figure}

To determine whether a vehicle is located within the ROI, we adopt the method described in \citep{da2024smart}. This approach compares the detection bounding box with a mask extracted from a parking lot image, where black pixels represent points inside the ROI and white pixels represent points outside. The ROI reference mask is shown in Figure~\ref{fig:roimask}.

\begin{figure}[htpb]
  \centering
  \includegraphics[width=0.49\textwidth]{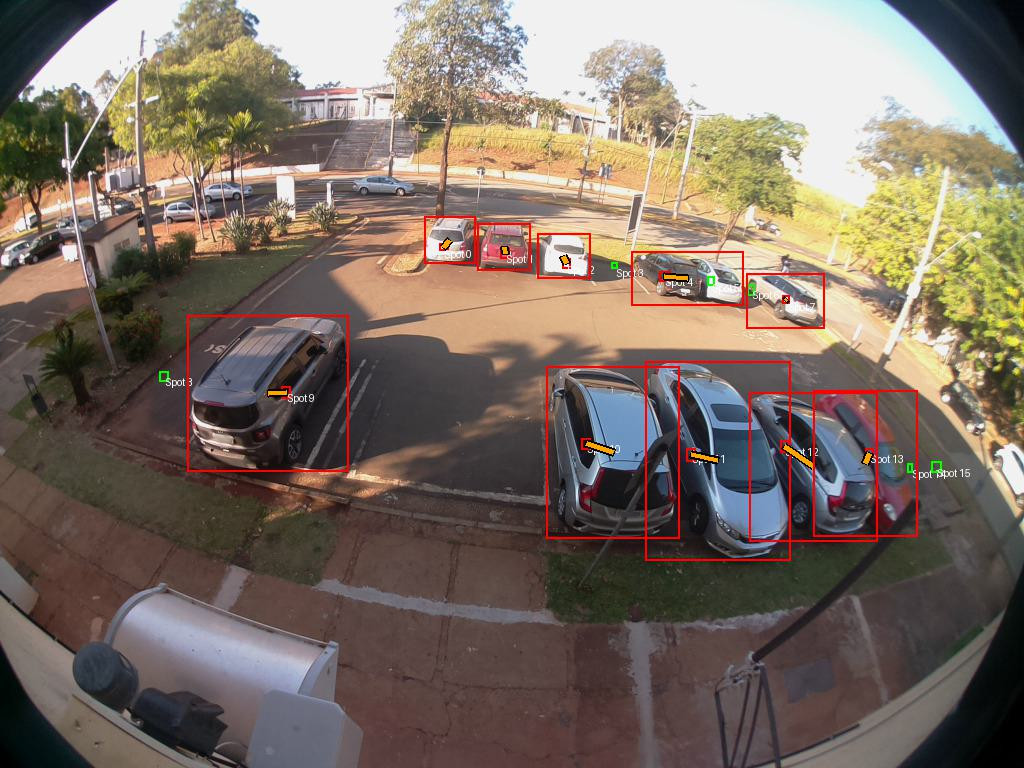} 
  \caption{Assigning method. \highlightyellowrevnew{The orange line indicate the distance of the center of the detection to the center of the labeled bounding box. The bigger red bounding boxes indicate detected vehicles, while the smaller bounding box is red for spots occupied and green for empty.}}
  \label{fig:assigning}
\end{figure}

Figure \ref{fig:assigning} shows the output of our method. Only cars located inside the parking lot are counted, and each is assigned to a predefined parking spot.

This simple method, which combines detection distance with the ROI, reduces false assignments produced by the deep learning models. Since each spot can be occupied by only one vehicle, cases where two bounding boxes overlap and are not eliminated during NMS and are counted as a single occupied spot.

\subsubsection{Adaptive Bounding Box Partitioning Method}
\label{subsec:abbp}

As illustrated in Figure~\ref{fig:assigning}, the YOLOv11m model occasionally merges multiple adjacent vehicles into a single bounding box. This happens, for example, in the upper-right corner, where two distinct vehicles are detected as one. To mitigate this issue, particularly in this region, we introduce the Adaptive Bounding Box Partitioning.

The method refines the set of detections based on the output of Algorithm \ref{alg:assign}. For this purpose, Algorithm \ref{alg:adaptive-split} verifies whether a detection overlaps with the critical spots in the top-right corner. If the bounding box area exceeds a predefined threshold, the box is vertically divided into two smaller bounding boxes, which are then incorporated into the existing detection set.

\begin{algorithm}[H]
\caption{Adaptive Bounding Box Partitioning}
\label{alg:adaptive-split}
\DontPrintSemicolon
\KwIn{Detections $B$, Spots $S$, image width $W$, image height $H$, critical spot IDs $C$, area threshold $A_{\text{th}}$, max distance $\delta$}
\KwOut{Refined detections $B'$}

$B' \gets \varnothing$\;
\ForEach{detection $d = (x_c,y_c,w,h) \in B$}{
    \ForEach{spot $s \in S$}{
        \If{$s.id \in C$}{
            $dist \gets \sqrt{(x_c - s.x_c)^2 + (y_c - s.y_c)^2}$\;
            \If{$dist < $$\delta$}{
                $area \gets (w \cdot W) \times (h \cdot H)$\;
                \If{$area > A_{\text{th}}$}{
                    $w' \gets w/2$\;
                    $B'.\text{append}((x_c - w'/2, y_c, w', h))$\;
                    $B'.\text{append}((x_c + w'/2, y_c, w', h))$\;
                    \textbf{continue with next detection}\;
                }
            }
        }
    }
    $B'.\text{append}(d)$ \tcp{Keep detection if not split}
}
\Return{$B'$}\;
\end{algorithm}

The area threshold was determined \hlsecondround{initially by} visual inspection of the test dataset\hlsecondround{, but was validated and refined through a statistical method based on the $Z-score$.}. Specifically, we \hlsecondround{first} analyzed the typical areas of detections in that region and selected a threshold corresponding to bounding boxes large enough to be confidently attributed to errors of the deep learning model. This was combined with an area distribution analysis based on histograms for each parking spot and the $Z-score$ method. With this analysis, described in detail in the Appendix, we discovered that an area of 5674 pixels would be large enough to cover more than one spot at that top right corner.\hlsecondround{ One limitation of this approach is that unusually large vehicles such as delivery vans may produce bounding boxes that exceed this threshold. However, in this parking lot is not common to have such larger vehicles parked there. Also, if the vehicle distribution or the camera angle changes, the threshold can be recomputed and updated from the statistical analysis.}

\begin{figure}[!h]
    \vspace*{-\topskip}
    \centering
    \begin{minipage}[b]{0.32\textwidth}
        \centering
        \includegraphics[width=\textwidth]{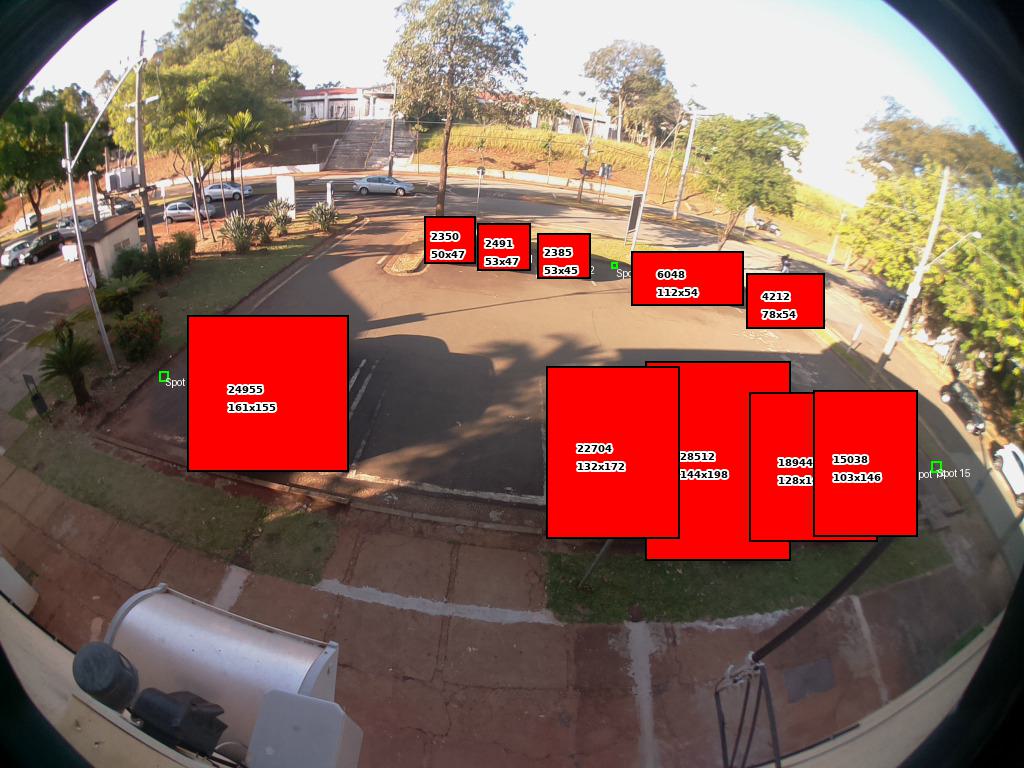}
    \end{minipage}
    \hfill
    \begin{minipage}[b]{0.32\textwidth}
        \centering
        \includegraphics[width=\textwidth]{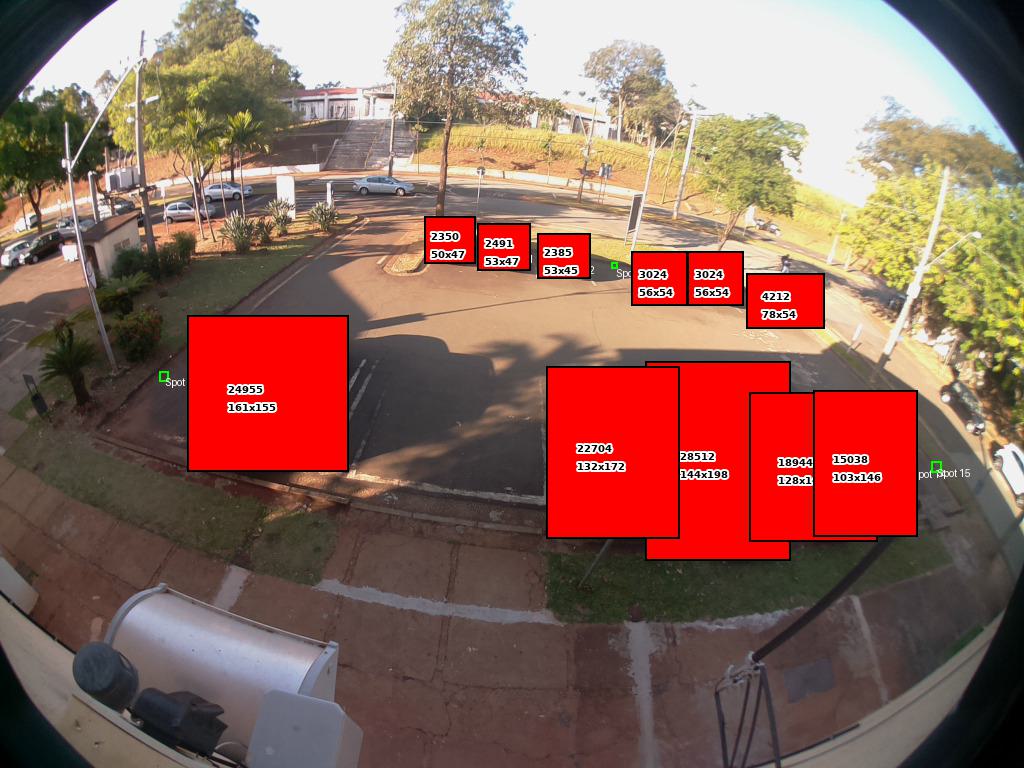}
    \end{minipage}
    \hfill
    \begin{minipage}[b]{0.32\textwidth}
        \centering
        \includegraphics[width=\textwidth]{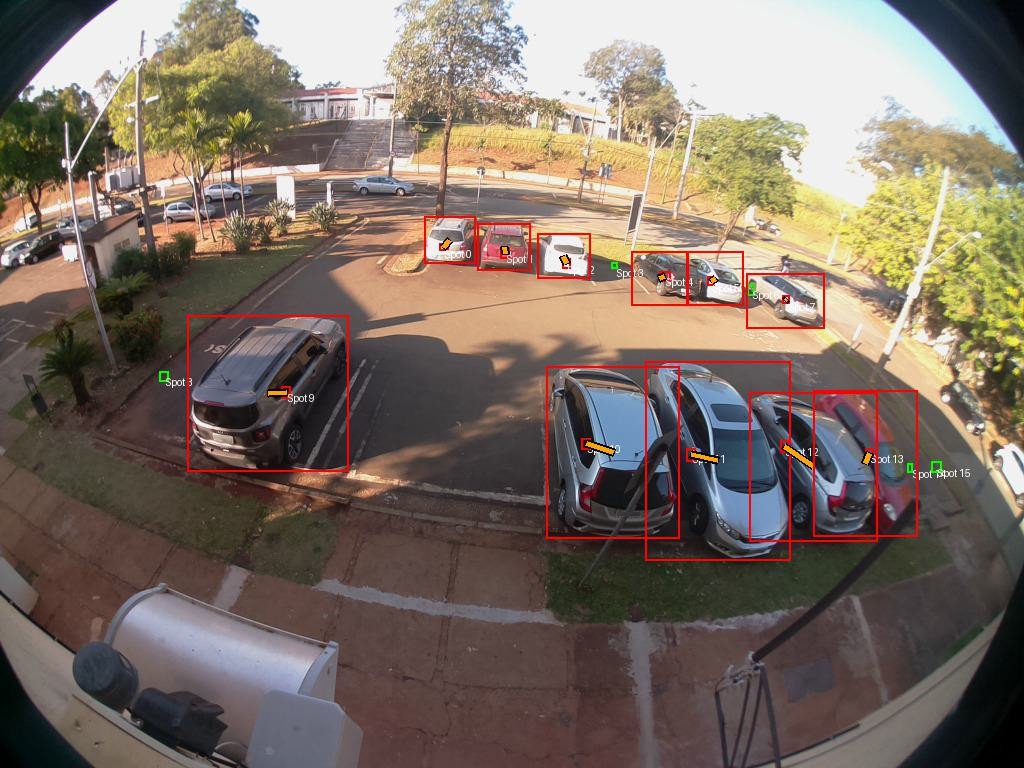}
    \end{minipage}

    \caption{Visual explanation of Adaptive Bounding Box Partitioning. 
    The area threshold was discovered by visually inspecting the dataset \hlsecondround{combined with a statistical method based on the $Z-score$}. 
    (a) Areas of detections that need refinement, (b) areas split according to threshold, 
    (c) final detection after applying ABBP.}
    \label{fig:adaptive_bb_partitioning}
\end{figure}

We can see the visual impact of applying the method at Figure \ref{fig:adaptive_bb_partitioning}. \highlightyellowrevnew{Note that first we show the areas of each detection followed by the output} of the method applied with the two new areas of 56$\times$54, that previously corresponded to a big bounding box of 112$\times$54 pixels. Finally, we can see the prediction and the assigned spots.  

\subsubsection{Compact Representation of Parking Occupancy}

To efficiently transmit parking occupancy data from the edge device to the InfluxDB time-series database, we encode the status of all parking spots in a compact binary representation. Let $S = \{s_1, s_2, \dots, s_{16}\}$ denote the set of parking spots, ordered by their IDs. Each spot $s_i$ is assigned a binary value:

\[
b_i =
\begin{cases}
1, & \text{if spot } s_i \text{ is occupied,} \\
0, & \text{if spot } s_i \text{ is free.}
\end{cases}
\]

The complete parking status is then represented as a binary string:

\[
\text{parking\_status} = b_1 b_2 \dots b_{16}.
\]

For example, a status of

\[
\text{parking\_status} = 1000011001100110
\]

indicates that 7 spots are occupied and 9 are free to park. To facilitate storage and transmission, this binary string is converted into a single integer value using:

\[
\text{parking\_bitmask} = \sum_{i=1}^{16} b_i \cdot 2^{16-i} = 34406
\]

This approach allows efficient storage of occupancy information while preserving the exact state of each spot and can be extended to any number of parking spots.

\subsection{Application Support Server}

\begin{figure}[!htpb]
  \centering
  \includegraphics[width=0.49\textwidth]{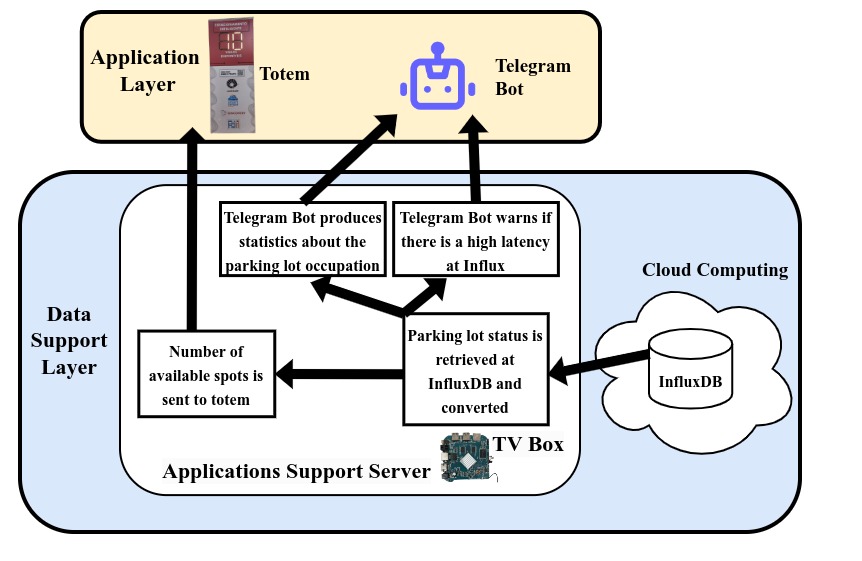} 
  \caption{Application Support Server Flow}
  \label{fig:flowtvbox}
\end{figure}

The application support server acts as an intermediary layer between the edge devices and cloud applications, 
aggregating data from InfluxDB and distributing it to  different applications such as the totem display and the Telegram Bot. 
This design improves reliability through latency monitoring and data parsing, enabling a solution that is easier to maintain, since any changes can be managed at the server without requiring modifications on the \hlsecondround{microcontroller} or low-level hardware of applications.

To host this service, we adopted a TV Box device, motivated by recent studies that demonstrated its suitability for IoT applications \citep{bannwart2025tv,da2024repurposing, sato2025people, courb,moreira2022agrolens,shu2021eight}. The TV Box employed in our experiments is the Youit TX2, equipped with an ARM Cortex-A7MP (32-bit) quad-core processor clocked at 1.2 GHz, 2 GB of LPDDR3 RAM, and running Armbian 25.8.0 Bookworm. Two main services are hosted by the server: the totem update system and the Telegram bot.

\subsubsection{Totem Update System}

The totem installed at the parking facility is directly updated by the support server. 
Rather than querying the database, it receives already processed and validated occupancy data. 
This reduces computational load, simplifies synchronization, and ensures consistent and reliable updates at the display. 
Furthermore, hosting this interface on the support server enables future extensions, such as remote monitoring of the totem’s status, automated health checks, and over-the-air firmware updates.

\subsubsection{Telegram Bot}

A Telegram bot queries the database and notifies if any delays happened in data transmission by comparing the current time with the timestamp of the last update. This is included as a monitoring mechanism for possible high latency at sending the data to InfluxDB. If the delay exceeds a configurable tolerance time, a notification is sent via the Telegram Bot API. Nevertheless, if the data is not sent to the database at the moment after the inference, it is still stored locally at Redis inside the edge device and sent after connection is reestablished.

Another feature introduced in this update is the generation of parking lot occupancy statistics. Daily and weekly reports are automatically sent, providing both textual summaries and visualizations. This functionality enables automated monitoring, fault detection, and reporting without requiring direct database access by end users. The notification system also assists in identifying outdated ROI masks or labeling tables of parking spot locations, which may occur if the camera position changes. Although this remains a limitation of our method, updating the system is straightforward, as new input files can be remotely uploaded to the edge device.

\subsection{Totem}

On the totem side, an ESP8266 periodically requests the latest data from the application support server and forwards it to the LED panel for visualization. Regarding the circuit, the previous schematic was preserved, with the addition of 1.2~k$\Omega$ resistors placed between the multiplexer and the MOSFETs.

\begin{figure}[H]
    \centering
    \includegraphics[width=0.49\textwidth]{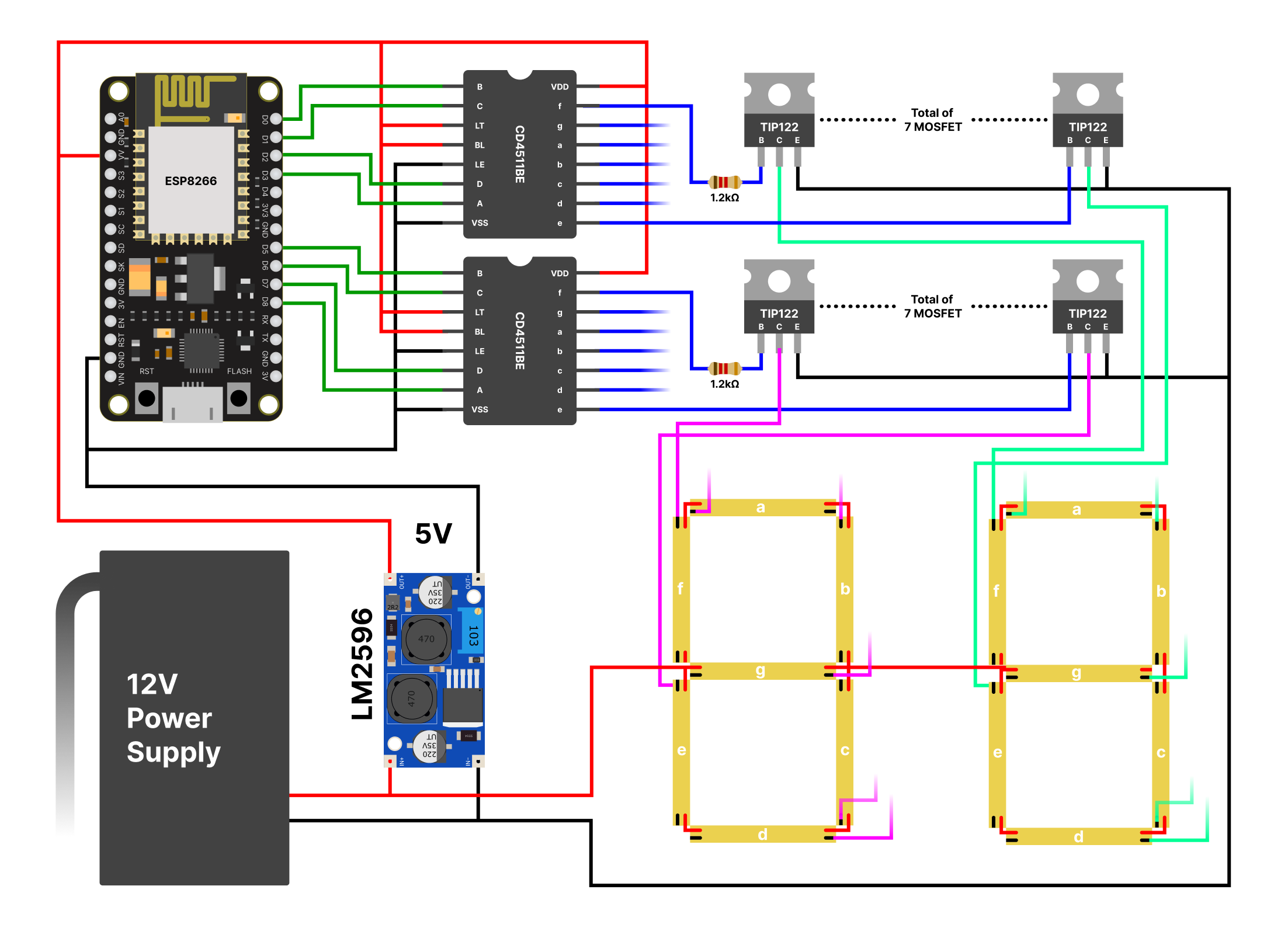}
    \caption{Schematic of how the totem works.}
    \label{fig:esq_totem}
\end{figure}

Figure \ref{fig:esq_totem} provides an overview of the totem circuit. It was built with two panels, one on each side, ensuring that the number of available spaces is visible from either direction. The microcontroller periodically retrieves the last data stored in InfluxDB and displays it on the panels. The entire circuit is powered by a 12V supply. However, since the development board and the multiplexers \footnote{\href{https://www.alldatasheet.com/datasheet-pdf/pdf/26904/TI/CD4511BE.html}{Multiplexer datasheet}}  operate at 5V, an LM2596 voltage regulator is used to step down the voltage. The power lines are indicated by the red and black tracks in Figure \ref{fig:esq_totem}.

To control each of the 14 segments of the panel individually, the specialized multiplexer CD4511BE was used. This component decodes its binary-coded decimal (BCD) input into the corresponding signals required for a 7-segment display. To display the correct digits, the development board configures its I/O pins to provide the appropriate BCD values required by the decoder, as illustrated by the dark green traces in Figure \ref{fig:esq_totem}. Since the multiplexer operates at a different voltage than the LEDs, TIP122 MOSFETs were used to switch the segments on and off. Additionally, to ensure visibility from both the front and back of the totem, the circuit was duplicated for the second panel. This duplication includes the MOSFETs and the two digits shown in the figure, effectively mirroring the entire section of the circuit. To prevent the multiplexer from outputting more current than it is rated for, a resistor was added before each MOSFET. 

\begin{figure}[htbp]
    \centering
    \begin{minipage}[c]{0.30\textwidth}
        \centering
        \begin{subfigure}[b]{\linewidth}
            \includegraphics[width=\linewidth]{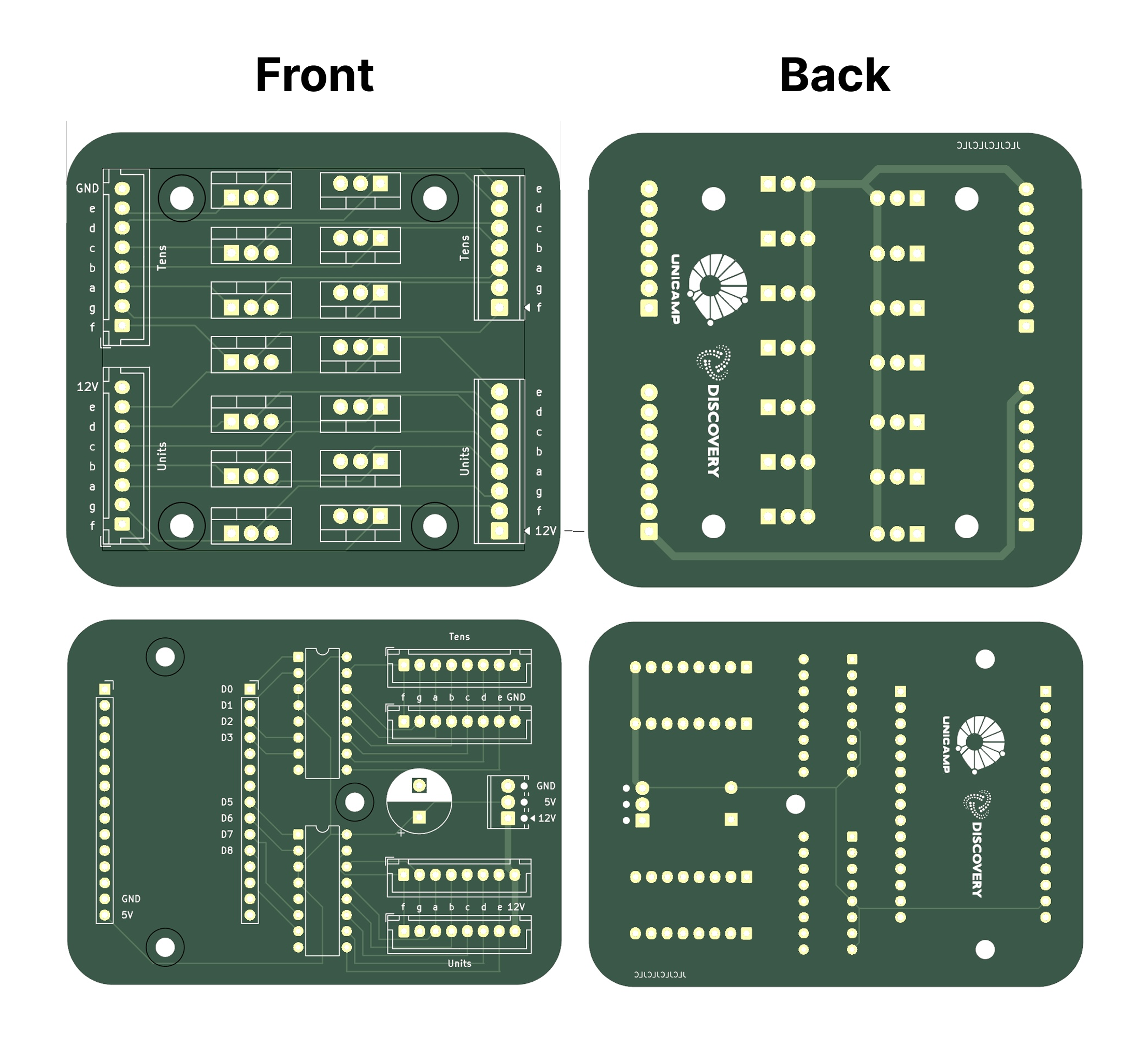}
            \caption{PCB render}
            \label{fig:pcb_render}
        \end{subfigure}
        
        \vspace{1em} 
        
        \begin{subfigure}[b]{\linewidth}
            \includegraphics[width=\linewidth]{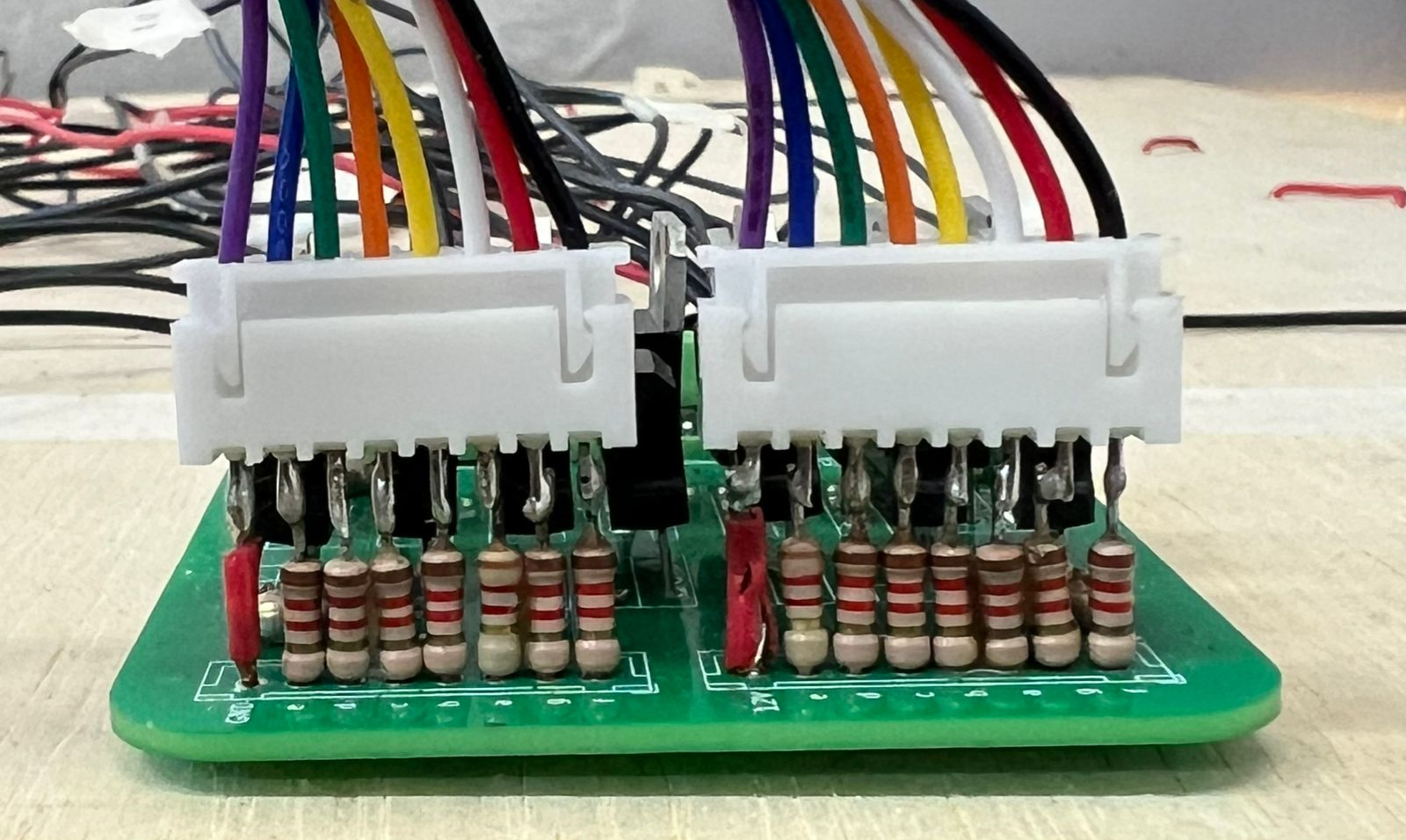}
            \caption{Side view of assembled PCB.}
            \label{fig:pcb_side}
        \end{subfigure}
    \end{minipage}
    \hfill 
    \begin{minipage}[c]{0.34\textwidth}
        \centering
        \begin{subfigure}[c]{\linewidth}
            \includegraphics[width=\linewidth]{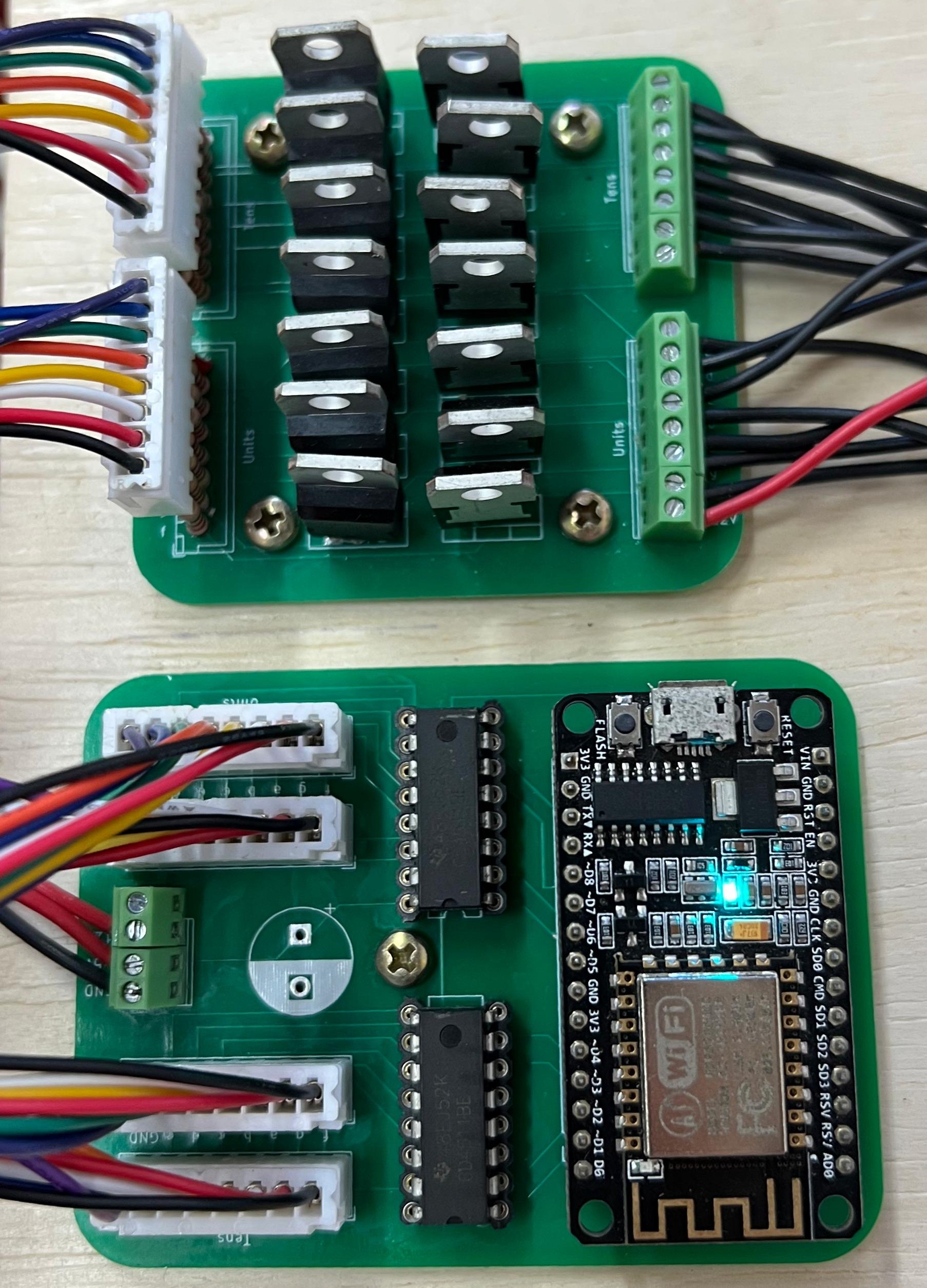}
            \caption{Top view of assembled PCB.}
            \label{fig:pcb_top}
        \end{subfigure}
    \end{minipage}
    \hfill
    \begin{minipage}[c]{0.21\textwidth}
        \centering
        \begin{subfigure}[c]{\linewidth}
            \includegraphics[width=\linewidth]{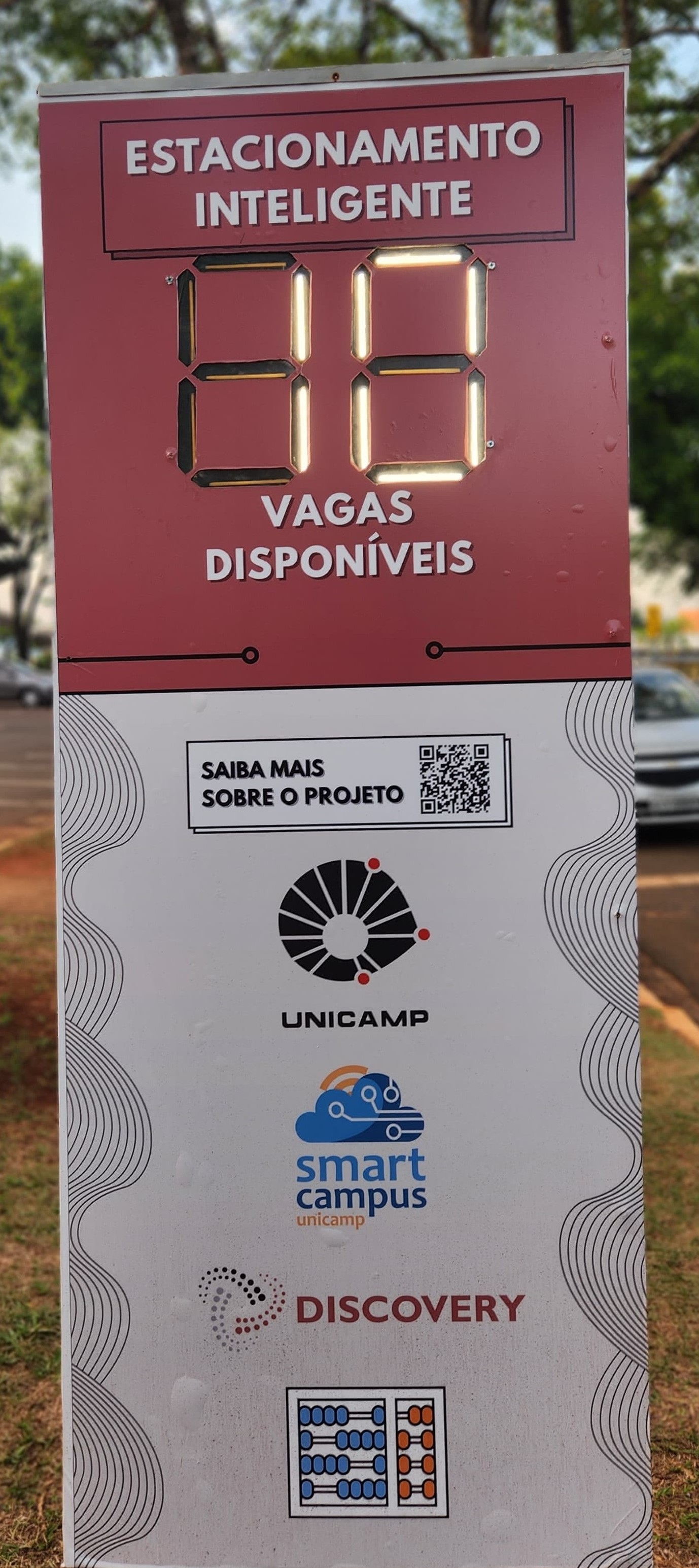}
            \caption{Circuit installed at the parking lot totem.}
            \label{fig:totem}
        \end{subfigure}
    \end{minipage}

    \caption{PCB assembly and installation details.}
    \label{fig:pcb_all}
\end{figure}

In order to ensure that only up-to-date and valid values are presented, a set of validation checks was implemented. If the retrieved value is negative or greater than the number of spots, the display is turned off. The same procedure is applied if no successful request is made within 5 minutes.

We also designed PCBs from the schematic, had them professionally manufactured, and soldered them, as shown in Figure \ref{fig:pcb_all} (a). Compared to the previous protoboard implementation, this approach provides a clearer layout, reduced wiring complexity, improved readability, and greater mechanical and electrical reliability. It also allows consistent replication of identical boards, enhancing scalability and long-term durability. Due to a design oversight, the PCBs lack dedicated footprints for the resistors, which were therefore mounted under the connectors, as shown in Figure~\ref{fig:pcb_all}~(b). The final circuit installed within the parking lot facilities is also illustrated in Figure~\ref{fig:pcb_all}~(d).

\subsection{Digital Shadow towards a Digital Twin}
The existing IC2 parking monitoring system provides a solid foundation \hlsecondround{towards the development of a} Smart Campus Parking Digital Twin. Conceptually, a Digital Twin can be described as comprising three main blocks: (i) the physical asset, (ii) the virtual counterpart (that reproduces the geometries, physics behavior, status, surrounding environment, etc.), and (iii) the communication medium  between them \citep{mihai2022digital}. It is essential to note that a \hlsecondround{fully realized Digital Twin} is not limited to \hlsecondround{data visualization or state representation; it also encompasses semantics and the ability to reproduce system behavior under hypothetical conditions, that is predictive or exploratory simulation capabilities and a bidirectional flow where the digital entity can influence or inform the physical system. 

At the current stage, the IC2 system implements a Digital Shadow, characterized by a one-way real-time data flow from the physical infrastructure to the virtual representation. This provides an interoperable and standardized visualization environment but does not yet incorporate the simulation, feedback, or decision-support mechanisms that would complete the Digital Twin loop}. As highlighted by Piroumian et al.\citep{piroumian2023making}, the “\textit{true value of digital twins is in the potential achievement of interoperability of applications and application areas across technical domains, industries, and vertical markets},” \hlsecondround{a goal towards which the present work is designed to evolve.}

Achieving a higher degree of interoperability requires the adoption of standardized data exchange mechanisms, domain-specific data models, and common frameworks. In addition, realizing the full potential of \hlsecondround{a future Digital Twin will involve integrating dedicated functionalities, such as simulation, hypothetical scenario exploration, and AI-driven feedback mechanisms. In this sense, the proposed architecture is best understood as a one that structures the planned evolution towards a full Smart Campus Parking Digital Twin around the following four main components.}

\subsubsection{Component 1: Adoption of NGSI-LD as standardized data exchange protocol}
NGSI-LD (Next Generation Service Interfaces - Linked Data) was developed by the European Telecommunications Standards Institute (ETSI)\footnote{\url{https://www.etsi.org}} and defines an information model for context data in a standardized and interoperable way (based on JSON-LD\footnote{JSON for Linked Data, allows to add semantics to JSON data by using context definitions to map JSON keys to URIs - \url{https://json-ld.org}}  format) as well as an API to interact with this context data. It can be defined as an open framework for the exchange of contextual information for smart services, using a RESTful API that aims to make it easier to find and exchange information with open databases, mobile apps, and IoT platforms~\citep{bees2019ngsi}.

\subsubsection{Component 2: Data Modeling of Smart Campus Parking entities}
The Smart Campus Parking system is modeled using Smart Data Models\footnote{\url{https://github.com/smart-data-models}}, which provide a tailored implementation of domain-specific ontologies derived from the NGSI-LD information model. These models define standardized entity types, relationships, and properties in domains such as Smart Cities, ensuring semantic consistency, interoperability, and reusability across different systems and platforms.

\begin{figure}[htpb]
  \centering
  \includegraphics[width=0.49\textwidth]{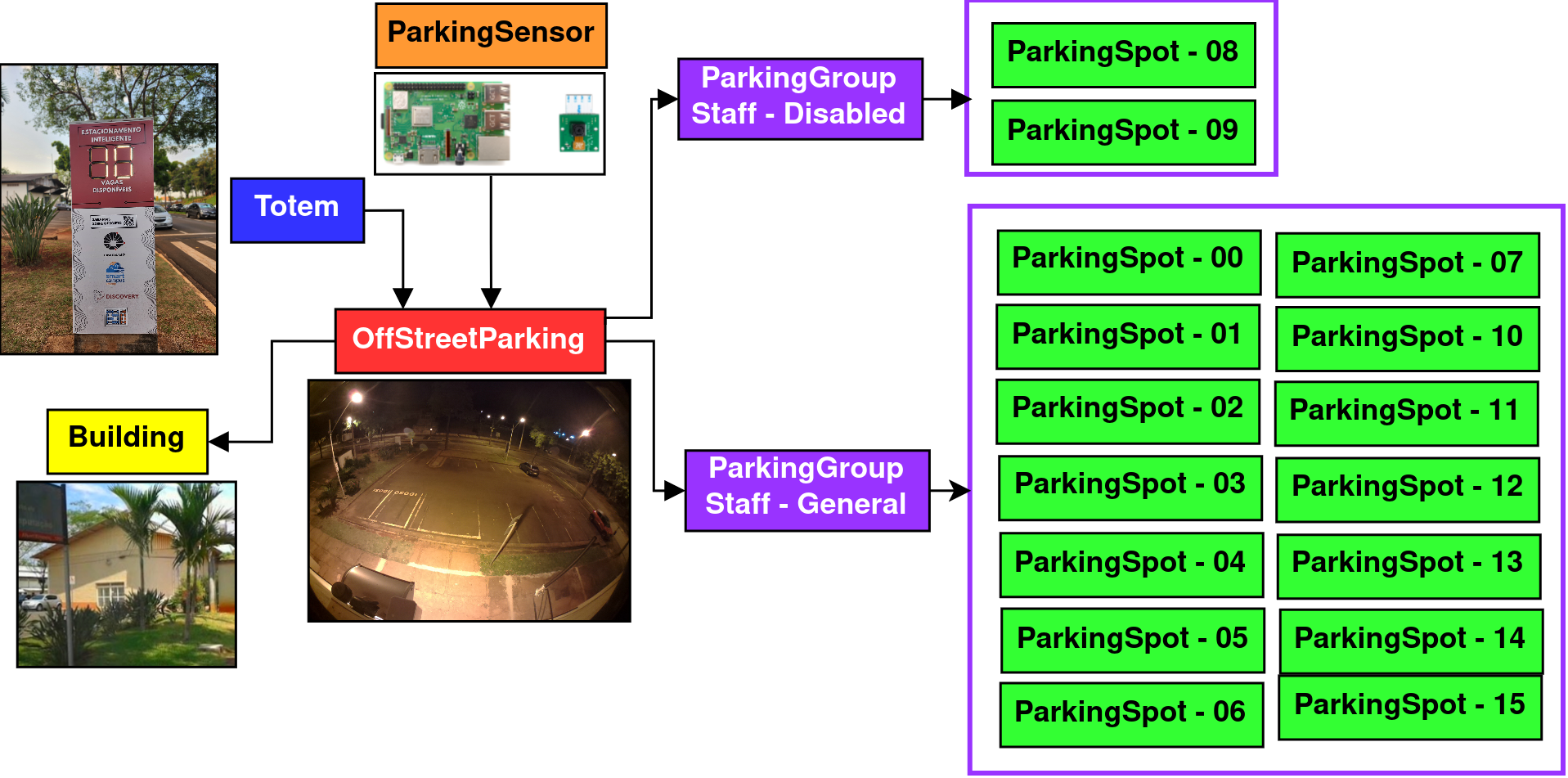} 
  \caption{Parking entities modeled using Smart Data Models}
  \label{fig:fiware-Modeling-scheme}
\end{figure}

As illustrated in Figure \ref{fig:fiware-Modeling-scheme}, six entity types were created to represent the parking environment: "\textit{Building}", "\textit{OffStreetParking}", "\textit{ParkingGroup}", "\textit{ParkingSpot}", "\textit{ParkingSensor}", and "\textit{Totem}".

Figure \ref{fig:fiware-Modeling-scheme} presents a simplified scheme of the entity modeling, highlighting their interconnections. For example, the "\textit{ParkingSensor}" captures images and computes the number of vehicles within the "\textit{OffStreetParking}" area. It also determines the occupancy status of each "\textit{ParkingSpot}", enabling the calculation of available spaces within different "\textit{ParkingGroup}"s - one for general staff and another for disabled staff. This aggregated availability information is then displayed to drivers through the "\textit{Totem}". The "\textit{Building}" entity serves as a reference point, associating the parking system with the campus infrastructure. Furthermore, higher-level entities can be added to group parking facilities under larger organizational structures, such as faculties, institutions, labs, libraries, or even the university as a whole.

\begin{figure}[htpb]
  \centering
  \includegraphics[width=0.49\textwidth]{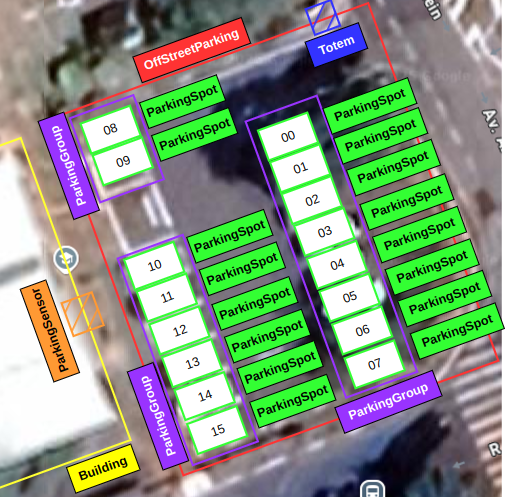} 
  \caption{Parking entities mapped to the parking area}
  \label{fig:fiware-parking-model}
\end{figure}

In Figure \ref{fig:fiware-parking-model}, the modeled entities are mapped to the IC2 Parking area. The "\textit{Building}" entity (yellow box) corresponds to the physical structure located in front of the parking area. The "\textit{OffStreetParking}" entity (red box) represents the entire \hlsecondround{parking} facility, including its associated "\textit{ParkingGroup}"s (purple boxes) and "\textit{ParkingSpot}"s (green boxes). The "\textit{ParkingSensor}" entity (orange box) models the Raspberry Pi with a camera, installed inside a protective case in the building. Finally, the "\textit{Totem}" entity (blue box) displays the real-time availability of parking spaces to incoming drivers.

\begin{figure}[htpb]
  \centering
  \includegraphics[width=0.49\textwidth]{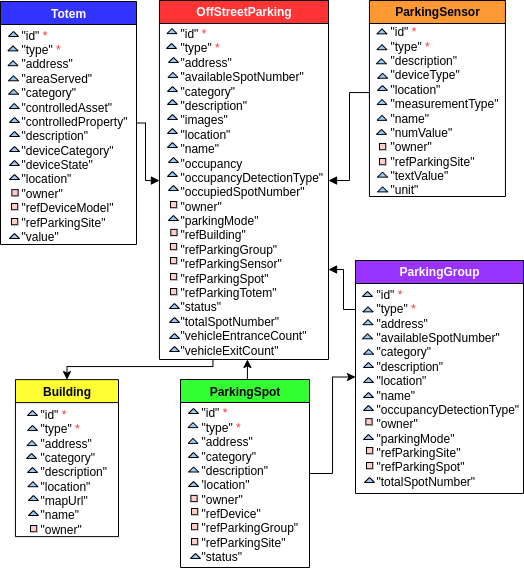} 
  \caption{Entity-Relationship diagram of the parking using Smart Data Models}
  \label{fig:fiware-Modeling}
\end{figure}

Figure \ref{fig:fiware-Modeling} provides a detailed view of each entity, where properties are represented by $\blacktriangle$ and relationships by $\blacksquare$. The fields "id" and "type" are mandatory (marked with *), although additional properties can also be defined as "required" depending on the modeling needs, providing flexibility in the design. 
Both the properties and relationships follow standardized definitions and specifications, which facilitates their reuse across entities and supports the creation of new ones based on these well-established elements.

Properties can generally be grouped into two categories. The first corresponds to static properties, such as "address", "category", "description", or "name", which describe structural aspects of the entity and rarely change. The second corresponds to dynamic properties, such as "availableSpotNumber", "occupiedSpotNumber", or "occupancy", which capture the current state of the entity and are updated more frequently. This distinction allows the model to simultaneously represent both the persistent characteristics of the system and its real-time operational conditions.

Relationships, on the other hand, are expressed as references to other entities through their ids. This establishes explicit links between entities. For example, each "ParkingSpot" references a "ParkingGroup", indicating that it belongs to a specific group. Likewise, each "ParkingGroup" references the "OffStreetParking" \hlsecondround{entity} to which it belongs, while both the "ParkingSensor" and the "Totem" reference the "OffStreetParking" \hlsecondround{entity} where they operate. Finally, the "OffStreetParking" \hlsecondround{entity} references the associated "Building".

The practical value of these relationships lies in the contextualization they provide: each entity becomes part of a broader semantic network. This enables the aggregation of information at higher levels (e.g., from parking spots to groups and entire facilities), supports scalability by allowing new entities to be added without redesigning the schema, and facilitates semantic queries and reasoning. For instance, a query such as "all available parking spots near Building X" can be resolved because the relationships connect buildings to parking areas, and parking areas to individual spots. The complete modeling, including specific values, attributes, and detailed description of each property and relationship, can be further explored in our repository\footnote{\url{https://github.com/discovery-unicamp/smartparking_unicamp/tree/main/digital_twin}}.

\subsubsection{Component 3: Integration of FIWARE software components}
FIWARE is a framework of open source software platform components\footnote{FIWARE Catalogue \url{https://www.fiware.org/catalogue/}} for building smart solutions, using a common API centered on Context Data Management across multiple domains. These components, referred to as \textit{Generic Enablers}, are built around the NGSI-LD information model and API, ensuring standardized information exchange and enabling interoperability.

FIWARE components provide the middleware that enables context-aware data management, interoperability, and integration across heterogeneous systems. 

\begin{figure}[htpb]
  \centering
  \includegraphics[width=0.9\textwidth]{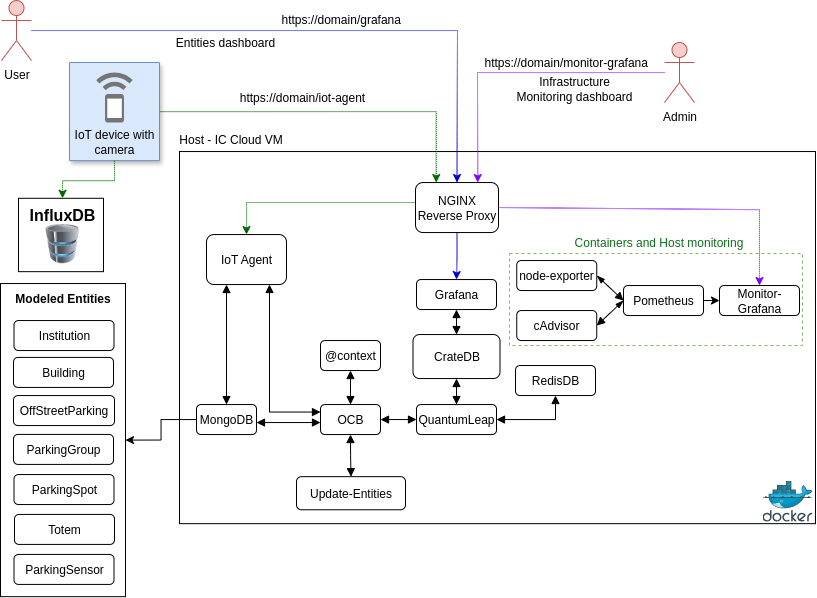} 
  \caption{Smart Campus Parking \hlsecondround{Digital Shadow} base architecture based on FIWARE components \hlsecondround{to evolve to a full Digital Twin}.}
  \label{fig:fiware-arch}
\end{figure}

The components of the architecture, as illustrated in Figure \ref{fig:fiware-arch}, are orchestrated using Docker Compose\footnote{https://docs.docker.com/compose/} and are described as follows.

  \textbf{NGINX\footnote{\url{https://nginx.org}}:} It converts external HTTP traffic to HTTPS for secure communication and serves as a single entry point to simplify access management. It also secures communication with external services using an SSL certificate, manages internal communication between containers over HTTP, and routes traffic to the IoT Agent and Grafana containers according to sub-paths. In addition, NGINX is optimized to handle many simultaneous connections efficiently.
  
  \textbf{Orion Context Broker (OCB)\footnote{\url{https://github.com/FIWARE/context.Orion-LD/tree/develop/doc/manuals-ld}}:} is a core FIWARE component that manages context information in the system. It follows the NGSI-LD Information Model and API, supporting entity CRUD operations and subscriptions. OCB stores the latest entity data in MongoDB and retrieves Smart Data Models from the @context container. It interacts with several components: receiving data from IoT sensors, sending commands to actuators through the IoT Agent for JSON, notifying QuantumLeap for storage in CrateDB and RedisDB, and updating entities through a FastAPI container. Since changes in one entity do not automatically update others, OCB requires explicit updates when values are interdependent.
  
  \textbf{@context:} this component runs an Apache\footnote{\url{https://httpd.apache.org/}} web server to serve the @context file, which defines the data models for Smart Parking entities. These models are customized for the UNICAMP Campus based on the Smart Data Models standard.
  
  \textbf{MongoDB\footnote{\url{https://www.mongodb.com/}}:} persistence layer for storing the current state of Smart Parking entities and configuration data.
  
  \textbf{IoT Agent for JSON\footnote{\url{https://fiware-iotagent-json.readthedocs.io/en/latest/}}:} it acts as a bridge between IoT devices and the Orion Context Broker (OCB). It supports multiple IoT protocols (such as LoRaWAN, Sigfox, and OPC-UA) and translates device messages into the NGSI-LD format. Devices send simplified messages with a device key, and the IoT Agent builds a complete request for the OCB, including the data model and entity ID.
  
  \textbf{QuantumLeap\footnote{\url{https://quantumleap.readthedocs.io/en/latest/}}:} subscribes to notifications from the Orion Context Broker (OCB), usually receiving sensor data. Unlike the OCB, which only stores the latest entity state in MongoDB, QuantumLeap converts NGSI-LD data into a time-series format for storage in CrateDB and RedisDB.
  
  \textbf{CrateDB\footnote{\url{https://cratedb.com/}}:} is a real-time analytics database optimized for time-series data, such as IoT sensor measurements. It supports SQL queries and includes features tailored for time-series analysis.
  
  \textbf{RedisDB\footnote{\url{https://redis.io/}}:} is an in-memory database used to cache data. 
  
  \textbf{Node Exporter\footnote{\url{https://github.com/prometheus/node_exporter}}:} an open-source monitoring tool that collects detailed host system metrics, including CPU, memory, disk, network, and other performance indicators.
  
  \textbf{cAdvisor\footnote{\url{https://github.com/google/cadvisor}}:} is a monitoring tool that tracks resource usage and performance of running containers.
  
  \textbf{Prometheus\footnote{\url{https://prometheus.io/}}:} is an open-source system monitoring and alerting toolkit, it collects and stores metrics from Node Exporter and cAdvisor as time series data.

  \textbf{Grafana\footnote{\url{https://grafana.com/}}:} is an open-source analytics and visualization platform for monitoring and observability. In this system, it displays historical entity data from CrateDB at the endpoint \textit{/grafana} and metrics of the host (from Node Exporter) and the containers (from cAdvisor) through Prometheus at the endpoint \textit{/monitor-grafana}.
  
  \textbf{Update-Entities container}: runs FastAPI\footnote{https://fastapi.tiangolo.com/}, which listens for notifications sent by the OCB about changes in the "\textit{ParkingSensor}" entity. So, when it receives the new status of the parking, it parses the data and calculate the new values for the "\textit{OffStreetParking}", "\textit{ParkingGroup}", "\textit{ParkingSpot}" and "\textit{Totem}" entities.

After introducing the individual components, it is essential to describe how they interact to deliver the system's functionality.

The process begins with the IoT device (a Raspberry Pi equipped with a camera) capturing images of the parking area and performing on-device inference. From this analysis, the device determines the status of each parking spot and encodes the result as an integer value. This payload is then transmitted via an HTTP request to the system endpoint (/iot-agent) as a JSON object, with the "parking\_status" field storing the integer value. Optionally, the data can also be sent to InfluxDB as a backup during the system’s deployment. The request includes the parameter "k" as the API key for authentication and "i" as the device identifier.

\begin{verbatim}
[language=bash]
curl -L -X POST \
    'https://<domain>/<endpoint>
    ?k=<api_key>&i=<device_ID>' \
    -H 'Content-Type: application/json' \
    --data-raw '{ "parking_status": 34406 }'
\end{verbatim}

The request is first received by the NGINX Reverse Proxy, which secures the communication with SSL certificates and routes the traffic to the IoT Agent. The IoT Agent then translates simplified device message into NGSI-LD format and updates the corresponding "\textit{ParkingSensor}" entity in the Orion Context Broker (OCB).

Upon receiving the update, the OCB updates the  "\textit{ParkingSensor}" entity and triggers a previously configured notification to the \textit{Update-Entities} container with the new "parking\_status" value. The \textit{Update-Entities} component decodes the integer payload into a binary string that represents the occupancy status of multiple spots (with 1 indicating occupied and 0 indicating free). It then computes aggregate values, such as the total number of available  and occupied spaces, availability per parking group, and the status of each individual spot. Using this information, it updates higher-level entities such as  "\textit{OffStreetParking}", "\textit{ParkingGroup}", "\textit{ParkingSpot}", and "\textit{Totem}" by sending the appropriate requests back to the OCB.

Simultaneously, the OCB forwards notifications to QuantumLeap, which persists selected entity attributes in CrateDB for historical analysis. This data is later visualized in Grafana dashboards, which are accessible to users through the /grafana endpoint managed by NGINX.

At the infrastructure level, system monitoring is handled by Node Exporter and cAdvisor, which collect host- and container-level metrics. Prometheus aggregates these metrics, while a dedicated Grafana instance, accessible via the /monitor-grafana endpoint, provides administrators with insights into the performance and reliability of the deployment.

\subsubsection{Component 4: Extension of the Digital Shadow software architecture to incorporate Digital Twin functionalities}
Evolving the current \hlsecondround{Digital Shadow parking} monitoring system, \hlsecondround{which offers a one-way, real-time representation of the digital asset,} \hlsecondround{towards} a full Digital Twin requires extending the software architecture. 
Two complementary capabilities \hlsecondround{guide this evolution}. The first is \hlsecondround{the} visual representation layer, whose detail may range from \hlsecondround{a} simplified 2D views to more elaborate 3D models, depending on \hlsecondround{system} requirements. \hlsecondround{In the current stage,} dashboards play a key role, offering intuitive real-time views of system data and states, and enabling end users to interact with the \hlsecondround{digital shadow} through clear and accessible interfaces.
A detailed description of these dashboards will be presented in Section \ref{sec:avaliacao}.
The second capability \hlsecondround{-still under development-} is the integration of \hlsecondround{simulation and agent-based reasoning mechanisms}. \hlsecondround{These functionalities are essential to move beyond visualization and towards a Digital Twin capable of running predictive analyses, testing hypothetical scenarios, supporting feedback into the physical system, and helping with decision-making.} Although the technical implementation of the AI agent \hlsecondround{and simulation modules} will be mentioned in Section \ref{sec:conclusao}, it is important to highlight that such \hlsecondround{functionalities are} essential for a \hlsecondround{fully realized Digital Twin. Thus, the present Digital Shadow architecture should be understood as the basis towards a full Digital Twin integration, representing a foundational step in the broader evolution from a Digital Shadow to a complete Smart Campus Parking Digital Twin.}

\section{Results and Discussion}
\label{sec:avaliacao}

\subsection{Model performance}

To address this, we added two additional model configurations to the existing benchmark discussed in previous works \cite{da202510,da2024smart}, which already included 13 models evaluated in recent years by our research team. The dataset used is the same of previous works with 4477 images containing various scenarios including day, night, rain, sun, and other variations. As it has more classes of spots \hlsecondround{than} of vehicles, the test set was composed of 3484 images that has at least one car, and balanced accuracy and Mean Absolute Error (MAE) were used as the evaluation metrics.

The methodology applied was the same of previous works, resizing the image to the shape YOLOv11m model was trained (640x640), following the same post-processing method to select the ROI. In addition to that, the methods of spot-wise approach and Adaptive Bounding Box Partitioning (ABBP) were applied and compared.

\begin{table}[htpb]
  \footnotesize
  \small
  \centering
  \setlength{\tabcolsep}{4pt}
  \begin{threeparttable}
  \rowcolors{2}{gray!10}{white} 
    \begin{tabular}{@{}lcccc@{}}
      \toprule
      \textbf{Model} &\textbf{B.Acc. (\%)} & \textbf{MAE} & \textbf{Time (s)\tnote{1}} & \textbf{MB} \\
      \midrule
      YOLOv9e\tnote{2} & 99.68 &0.03 & 92 $\pm$ 9.3 & 117.5 \\
      YOLOv11x\tnote{2} & 99.46 & 0.07& 46 $\pm$ 7.4 & 114.6 \\
      YOLOv11m\tnote{3} & 99.31 & 0.07& 11 $\pm$ 2.3 & 40.7 \\ 
      \textbf{\makecell[l]{YOLOv11m (TFLite)\\Spot-Wise + ABBP}} & \textbf{\highlightyellowrevnew{98.80}} &\textbf{0.08} &\textbf{8.5 $\pm$ 0.11} & \textbf{40.5} \\
      \textbf{\makecell[l]{YOLOv11m (TFLite)\\Spot-Wise}} & \textbf{98.71} &\textbf{0.10}  &\textbf{8.4 $\pm$ 0.12} & \textbf{40.5} \\
      YOLOv11m (TFLite)\tnote{3} & 98.65 &0.10 & 8.0 $\pm$ 0.62 & 40.5 \\ 
      YOLOv8x\tnote{2} & 98.49 &0.11 & 28 $\pm$ 0.7 & 136.9 \\
      YOLOv10x\tnote{2} & 98.30 &0.12 & 24 $\pm$ 0.2 & 64.4 \\
      YOLOv10n\tnote{2} & 97.32& 0.30 & 2 $\pm$ 0.03 & 5.9 \\
      YOLOv11n\tnote{2} & 96.15& 0.42 & 2 $\pm$ 0.02 & 5.6 \\
      Mask R-CNN\tnote{3} & 94.95&0.43 & -- & 257.6 \\
      YOLOv8n\tnote{2} & 94.86& 0.40& 2 $\pm$ 0.06 & 6.5 \\
      EfficientDet-D2 Lite\tnote{3} & 89.54&1.30 & 1 $\pm$ 0.11 & 7.6 \\
      YOLOv9t\tnote{2} & 86.90& 0.63& 2 $\pm$ 0.02 & 5.0 \\
      YOLOv3\tnote{3} & 83.55& 0.64& 9 $\pm$ 0.25 & 248.0 \\
      \bottomrule
    \end{tabular}
    \caption{Model performance ranked by balanced accuracy. The configurations introduced in this work are highlighted.}
    \label{tab:modelperf_table}
    \begin{tablenotes}
      \footnotesize
      \item[1] Mask R-CNN inference time could not be measured due to 1 GB RAM limit of the Raspberry Pi 3B+.
      \item[2] Values from \cite{da2024smart}.
      \item[3] Values from \cite{da202510}.
    \end{tablenotes}
  \end{threeparttable}
\end{table}

The spot-wise approach and ABBP within YOLOv11m (TFLite) yielded slight improvements in accuracy compared to the base model, without significantly affecting inference time or model size, represented by MB on Table \ref{tab:modelperf_table}, since these methods are applied only at post-processing. While the overall performance did not change substantially, the spot-wise approach enabled new features and analysis. Adding the spot-wise approach increased inference time by 0.4 seconds and improved balanced accuracy by 0.06 percentage points. Combining spot-wise and ABBP increased inference time by 0.5 seconds, achieving a \highlightyellowrevnew{0.15} percentage-point gain and reaching \highlightyellowrevnew{98.80}\% balanced accuracy. Regarding MAE, the proposed variants remain among the top-performing models. The YOLOv11m (TFLite) spot-wise + ABBP configuration reaches an MAE of 0.08, improving over the base TFLite model (0.10) and approaching the error levels of the larger YOLOv11x model, despite using an architecture nearly three times smaller in terms of MB. It is worth noting that the parking lot system updates once per minute, so the observed inference time of 8 seconds remains well within acceptable limits.

Although the improvements may appear modest, the main objective of this work was to gather occupancy data for each individual spot while maintaining low inference time, which was achieved without compromising accuracy. A qualitative assessment further allows a visual comparison between the models.

\begin{figure}[!htpb]
  \centering
  \includegraphics[width=0.9\textwidth]{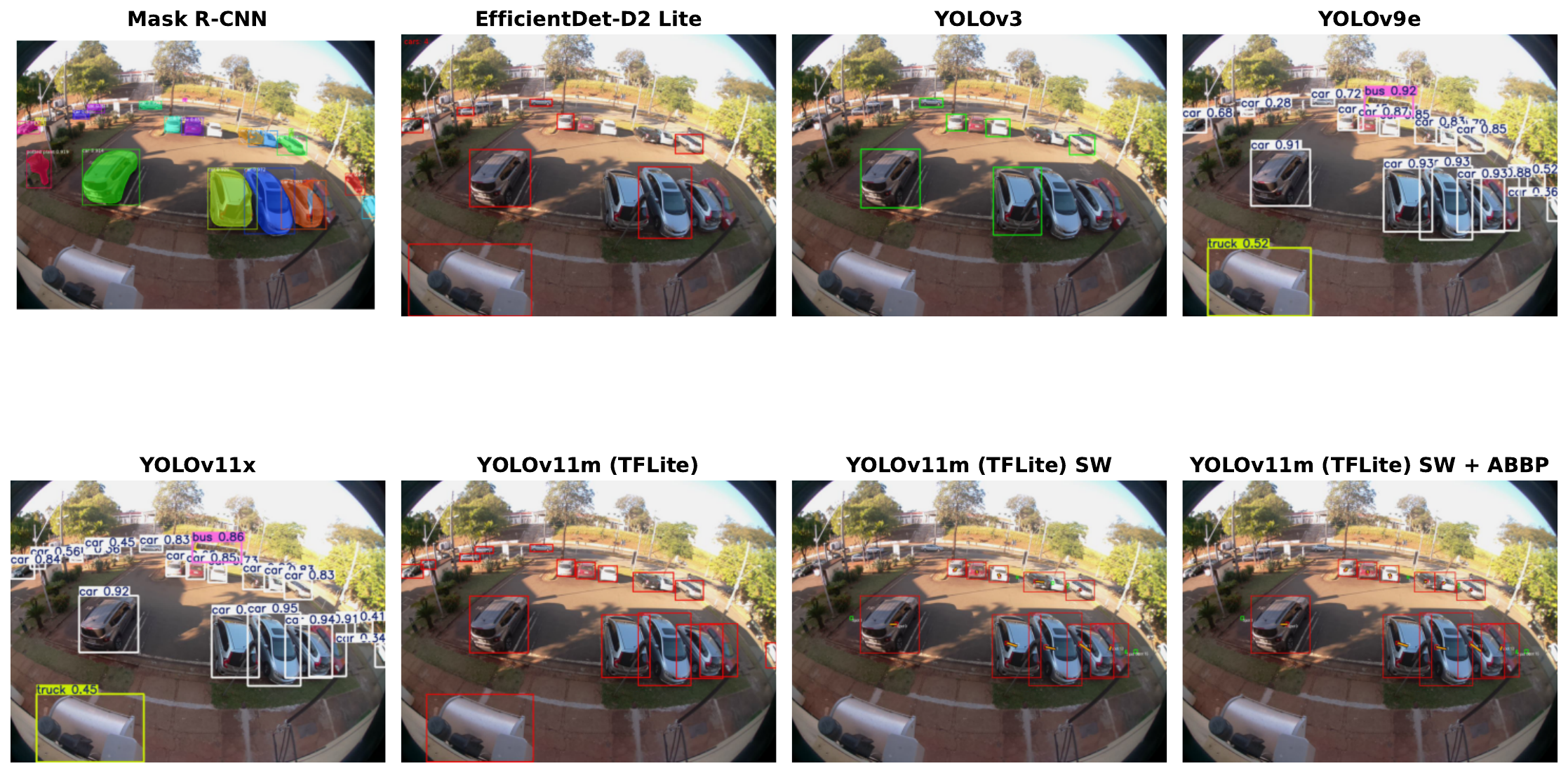} 
  \caption{Qualitative analysis of the evolution of the models throughout the years.}
  \label{fig:qualanalysis}
\end{figure}

Figure \ref{fig:qualanalysis} illustrates a challenging image with vehicles directly exposed to sunlight. The figure highlights how detection capacity has evolved over the years, reaching its highest performance with the larger YOLOv9e and YOLOv11x models. For deployment, however, we selected YOLOv11m (TFLite), which provides a lighter model with good balanced accuracy for the parking lot scenario. The figure also shows the impact of adding each proposed method: the spot-wise approach assigns each vehicle to its nearest parking spot, while ABBP improves performance in difficult cases, such as two adjacent cars that the base model initially classified as a single vehicle.

\subsection{System performance}

\subsubsection{Statistics about the parking lot occupation}

\begin{figure}[!htpb]
  \centering
  \includegraphics[width=0.7\textwidth]{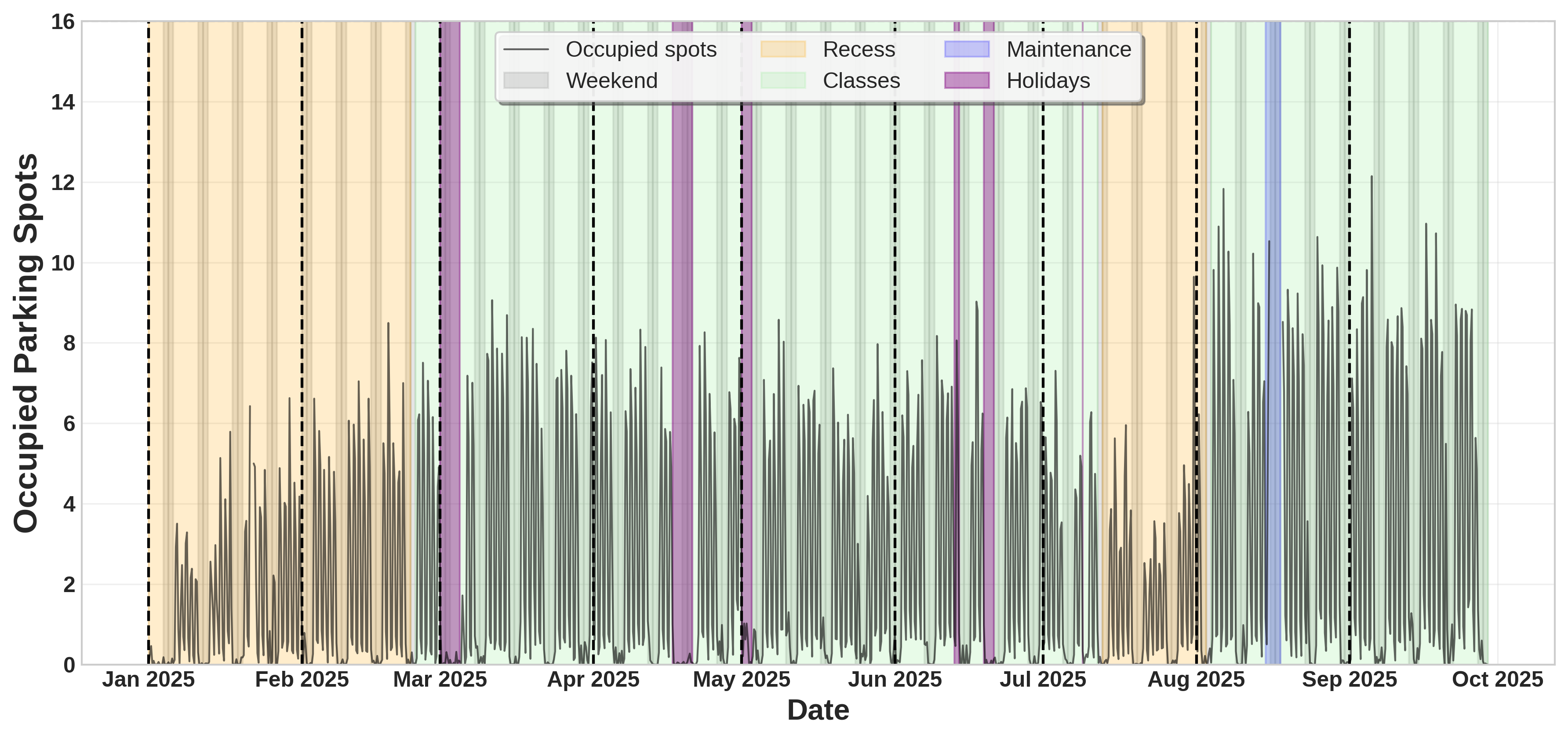} 
  \caption{Time series of occupied parking spots since January 2025. The green periods mark classes when the parking lot is more used than yellow periods that represent recess periods. Purple represents holidays, gray weekends, and blue represents a maintenance period where the parking status data representation changed. }
  \label{fig:influx}
\end{figure}

The Raspberry Pi responsible for vehicle detection was reinstalled in the parking lot in December 2024, with its output limited to the total number of cars and empty spots detected. In August 2025, the parking status representation was updated to also include the occupancy of each individual parking spot. In this subsection, we present and discuss the parking lot occupancy. The first analysis considers the time series of available spots since the system was installed.


Figure~\ref{fig:influx} presents real-world parking occupancy data collected during the first ten months of 2025, resampled to 6-hour intervals. The raw measurements were recorded by the Raspberry Pi and transmitted every minute to the InfluxDB server. The end-to-end latency, from image acquisition to visualization in the totem, is approximately one minute. This value was selected based on the average turnover of the parking lot, although the system could support shorter update intervals, since the inference step itself takes only 8 seconds.

A clear temporal pattern is notable analysing the data. During the period of March to July, the number of occupied spots remained consistently lower compared to the period of August to September. This variation correlates with the academic calendar: recess and holiday periods exhibit reduced demand, while occupancy increases during active class periods. The impact of weekends is also visible in the shaded areas of the timeline. 


\begin{table}[!htpb]
\centering
\tiny
\begin{tabular}{@{}lrrrrl@{}}
\toprule
\textbf{Spot} & \textbf{Hours} & \textbf{Hist. Avg.} & \textbf{Z-score} & \textbf{$\Delta$ Hours} & \textbf{Occupation} \\
\midrule
1  & 7.1  & 7.6 & -0.5 & -0.5 & Normal \\
2  & 1.4  & 4.7 & -2.2 & -3.3 & \textbf{Low Occupation} \\
3  & 8.3  & 4.2 &  2.0 & +4.1 & Normal \\
4  & 6.2  & 4.4 &  0.9 & +1.8 & Normal \\
5  & 4.8  & 3.9 &  0.4 & +0.9 & Normal \\
6  & 2.4  & 4.5 & -0.9 & -2.1 & Normal \\
7  & 4.5  & 4.5 & -0.0 & +0.0 & Normal \\
8  & 2.5  & 4.0 & -0.7 & -1.5 & Normal \\
9  & 2.7  & 3.7 & -0.4 & -1.0 & Normal \\
10 & 1.4  & 4.0 & -1.8 & -2.6 & Normal \\
11 & 10.8 & 6.9 &  2.9 & +3.9 & \textbf{Busy} \\
12 & 6.1  & 6.3 & -0.1 & -0.2 & Normal \\
13 & 4.7  & 5.4 & -0.3 & -0.7 & Normal \\
14 & 6.0  & 5.0 &  0.5 & +1.0 & Normal \\
15 & 1.9  & 3.3 & -1.1 & -1.4 & Normal \\
16 & 2.2  & 3.5 & -0.7 & -1.3 & Normal \\
\bottomrule
\end{tabular}
\caption{Daily occupation statistics for 2025-10-22. Each spot's daily occupied hours are compared to its historical weekday average using the $Z$-score. A anomalous spot is flagged as Busy or Low Occupation when $|Z| > 2$, indicating higher or lower use than usual, respectively. 
Overall statistics: Total occupied hours = 73; Average per spot = 4.6h; Most occupied = spot 11 (10.8h); Least occupied = spot 2 (1.4h); Spots with $<$ 1h = 0; Spots with $|Z|>2$ = 3.}
\label{tab:daily-stats}
\end{table}

These insights not only confirm the effectiveness of the proposed system for monitoring parking availability but also highlight its potential for broader applications. For instance, real-time knowledge of occupancy trends could support traffic management inside the campus, guide the allocation of additional parking areas, and provide valuable input for future smart mobility initiatives.

With the implementation of the spot-wise approach, additional insights could be incorporated into the analysis. The Telegram bot calculates daily occupancy for each spot as the total number of hours it remains occupied. As the InfluxDB used in this study has a temporal resolution of one minute, each row corresponds to an observation of all parking spots that are broken into spot columns. 
For each spot column, a value of $1$ indicates that the spot was occupied during that minute, while a value of $0$ indicates that it was free. 
Therefore, the total number of minutes that a given spot is occupied in a specific period can be obtained by summing all values equal to $1$. 
Dividing this result by 60 provides the total occupied hours. For each spot, the value of hours occupied in a day is compared with its historical mean and the number of absolute standard deviations ($Z-score$) per spot is observed, flagging a spot as anomalous when $|Z| > 2$. We ensured that this comparison is made with only weekdays when it is a weekday and only weekends when it is a weekend or holiday. This allows the system to tell if a spot is more or less occupied than its usual occupation. In addition, as shown in Table~\ref{tab:daily-stats} we reported daily overall statistics including the total and average occupied hours, the most and least occupied spots, and the number of spots with less than one hour of use.

\begin{figure}[H]
  \centering
  \includegraphics[width=0.44\textwidth]{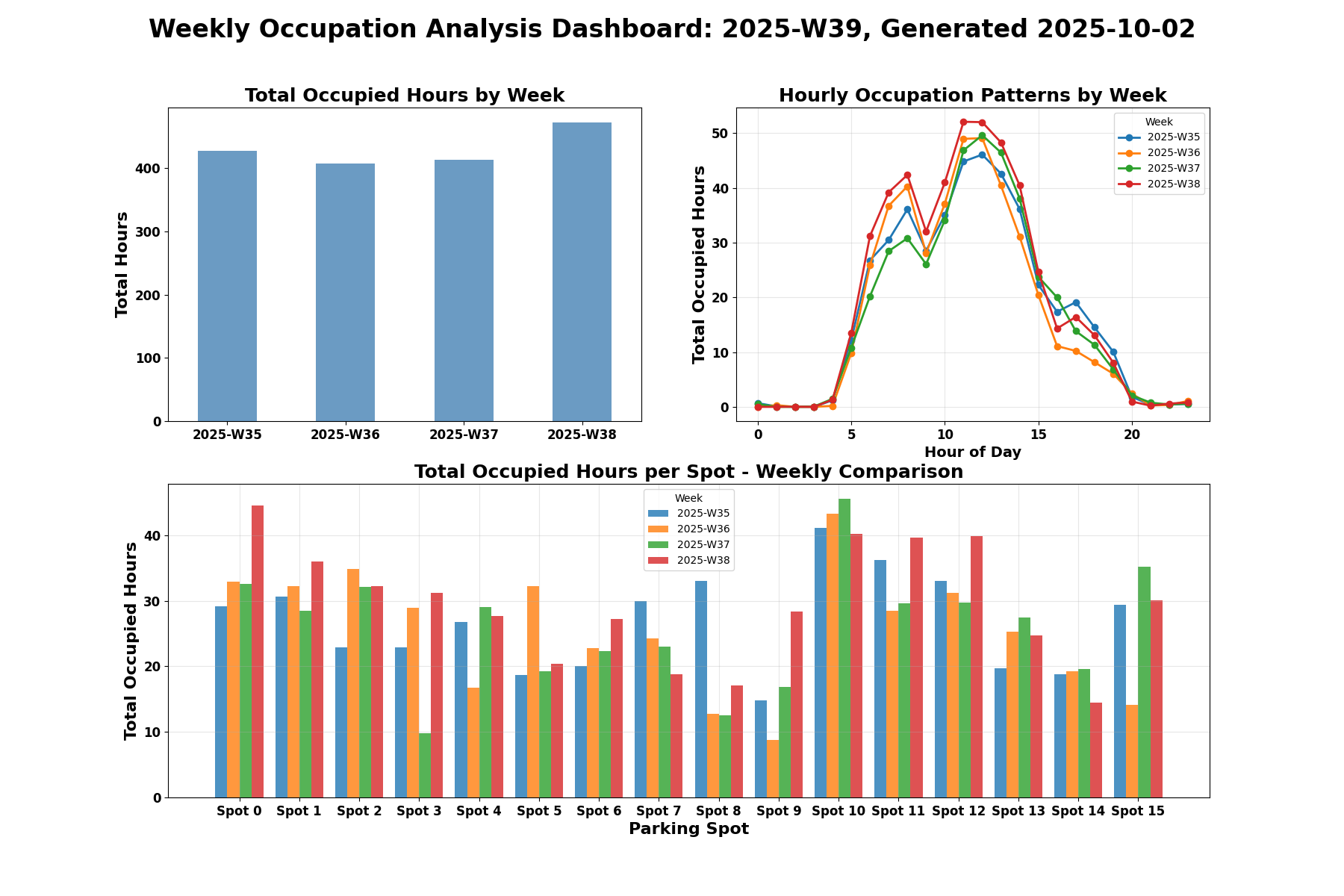} 
  \caption{Monitoring Dashboard with spot-wise approach}
  \label{fig:influx_spots}
\end{figure}
Figure \ref{fig:influx_spots} shows a configurable dashboard that contains the most relevant analysis to see if everything is normal with the system.

Based on the individual occupation of each spot, aggregating all hours of spots yields the overall occupation time of the parking lot in a week. 
It is possible to check if the total hours occupied in the parking lot are following the usual pattern, and also check the hourly patterns of occupation and total hours occupied per spot. The current pattern reveals a higher demand between 8:00 and 20:00, coinciding with class schedules and administrative activity at the university. This alignment with academic routines suggests that the system captures meaningful behavioral patterns of campus mobility. 
It is important to note we are measuring the cumulative time of occupation, not the number of distinct vehicles. 
For instance, a single vehicle parked for ten hours contributes the same total occupation as ten vehicles that each remain for one hour. 
Thus, this approach is particularly suited for evaluating the intensity of use.

\subsubsection{Digital Shadow Dashboards}

\begin{figure}[htpb]
    \centering
    \includegraphics[width=0.8\textwidth]{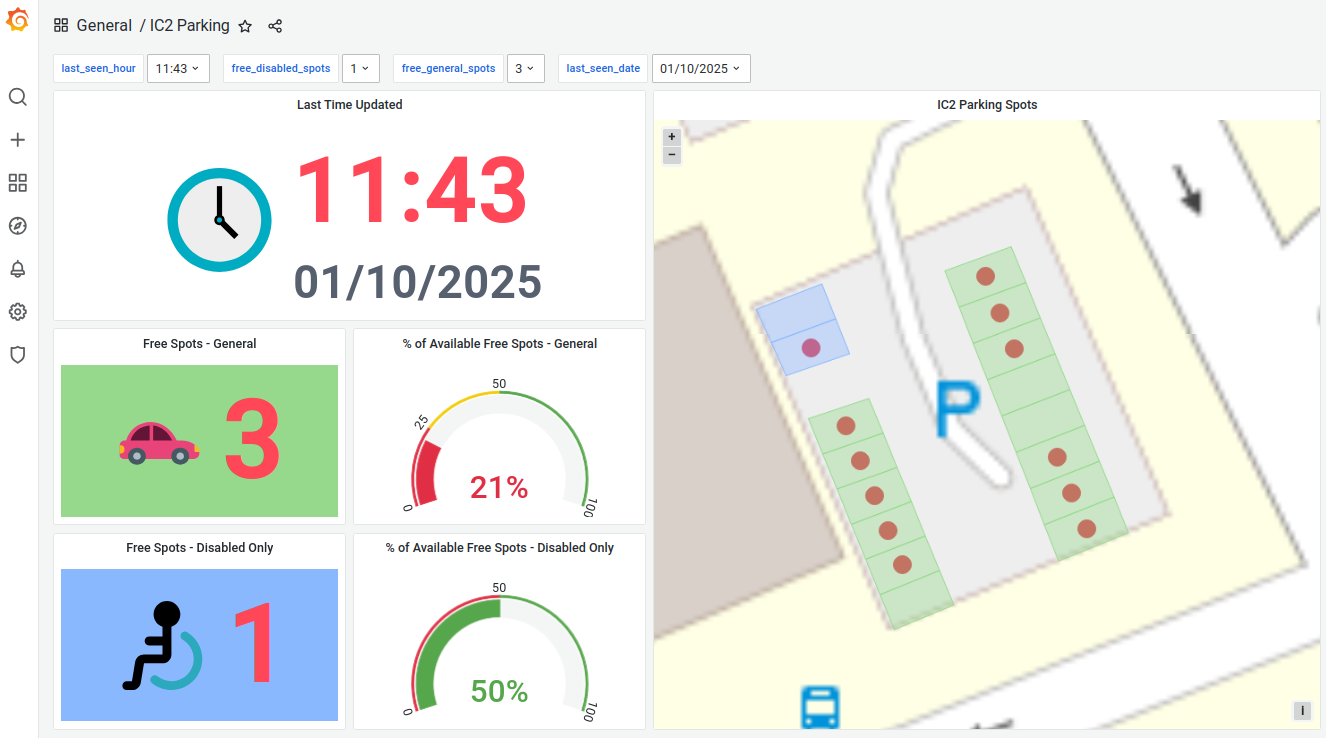}
    \caption{IC2 Parking UI for users}
    \label{fig:parking-grafana}
\end{figure}

Figure \ref{fig:parking-grafana} illustrates the user interface of the Digital Shadow, designed to provide an intuitive, real-time overview of parking availability and occupancy. The dashboard is organized into two main sections: a data summary panel on the left and a spatial visualization map on the right.

The data summary panel displays the timestamp of the most recent system update. It also presents key metrics for both user groups: for general staff, the number of available parking spots along with their proportion relative to the total in this category; and for disabled-only staff, the corresponding number of available spots and their proportion of the total.
Each metric is paired with a semantically meaningful icon (a standard car for general use, a wheelchair symbol for disabled access) and large, color-coded numerals to enhance readability.
The map represents a schematic layout of the parking area, with individual spaces represented as rectangles. These are color-coded according to user group: light green for general staff and sky blue for disabled-only spaces. Real-time occupied spots are indicated by small red dots within their corresponding rectangles. This visual encoding allows users to quickly identify free spaces without relying solely on numerical data.

\begin{figure}[htpb]
    \centering
    \includegraphics[width=0.44\textwidth]{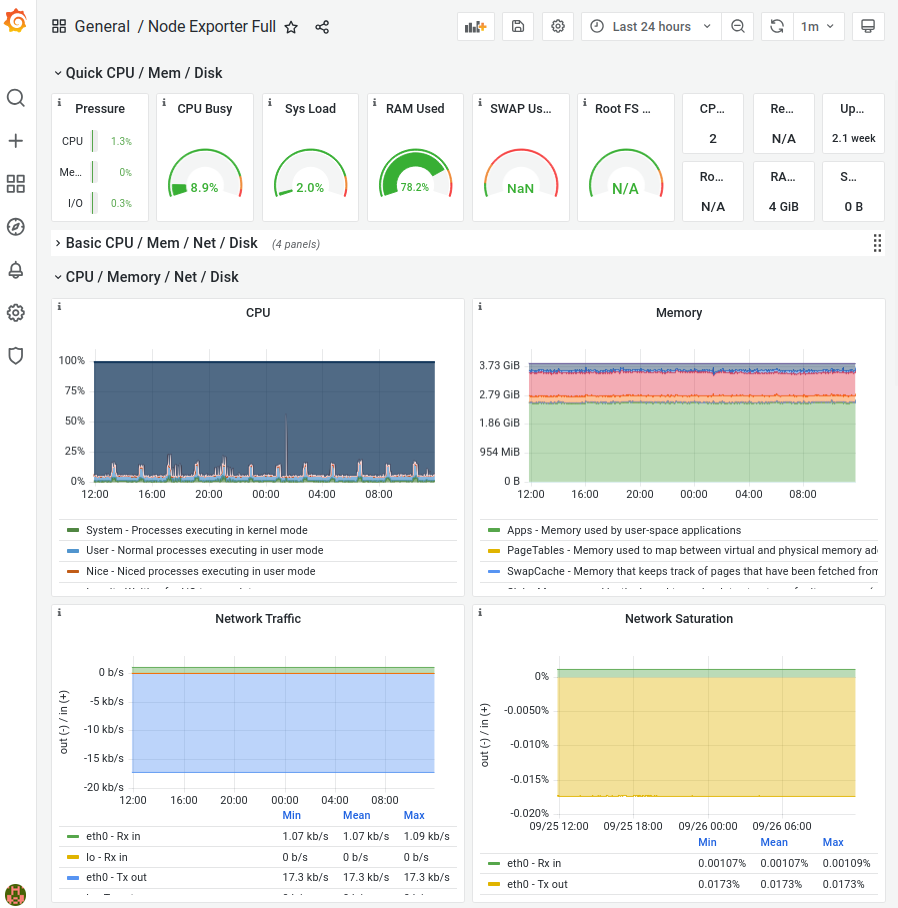}
    \caption{Node Exporter Metrics of Digital Shadow host machine}
    \label{fig:grafana-nodeexporter}
\end{figure}
Figure \ref{fig:grafana-nodeexporter}, extends the \hlsecondround{digital shadow} monitoring capabilities beyond parking data by incorporating system-level and container-level performance \hlsecondround{metrics}. This view integrates data collected from Node Exporter and cAdvisor, both exposed through Prometheus and visualized in Grafana, as shown in Figure \ref{fig:fiware-arch}.

Node Exporter provides detailed statistics about the host machine, including CPU utilization, memory consumption, disk I/O, and network throughput, thereby enabling administrators to assess the overall health and resource usage of the infrastructure supporting the digital twin. Complementarily, cAdvisor delivers fine-grained monitoring of containerized services, offering metrics such as per-container CPU and memory usage, filesystem activity, and network performance.

This way the system facilitates real-time performance tracking and long-term trend analysis.

\section{Conclusions and Future Work}
\label{sec:conclusao}
In this work, we introduced a spot-wise approach to an existing smart parking system at the Unicamp Institute of Computing, which previously estimated the number of available parking spots based solely on the total number of vehicles detected within the parking lot ROI. To implement this new approach, we incorporated two methods: a spot-wise strategy \highlightyellowrevnew{based on a distance-aware matching method with spatial tolerance} and an Adaptive Bounding Box Partitioning mechanism for challenging spots. This enhancement achieved a slightly higher balanced accuracy of \highlightyellowrevnew{98.80}\% while maintaining the same inference time of 8 seconds and a model size of 40.5 MB.

We also included two additional components in the system. The first is a \hlsecondround{foundational Digital Shadow of the IC2 Parking, aligned with} smart city standards to provide users and managers a visual representation of the digital entities present in the parking lot, enabled by the spot-wise approach. \hlsecondround{This is the initial step towards the evolution into a full Smart Campus Parking Digital Twin}. The second component is an application support server, hosted on a repurposed TV Box, which acts as an intermediate layer between the cloud database and the applications. This design simplifies system maintenance, as updates can be applied at the server level without requiring low-level modifications to the microcontrollers.

Beyond the Digital \hlsecondround{Shadow}, another application that benefited from the spot-wise approach is the monitoring bot, which can now generate daily and monthly statistical reports on parking spot occupancy. These statistics also serve to verify whether the system’s input files, namely the ROI mask of the parking lot and the table of annotated spot locations, remain properly configured. Changes in camera angle due to external factors such as weather may cause misconfigurations, requiring updates to these files. As future work, we plan to investigate automatic detection of camera angle shifts and implement corrective adjustments to the ROI mask and spot coordinates. An alternative approach is to integrate an inertial sensor into the edge device, coupled with a mechanism that ensures the camera remains stable.

While the smart parking system at the Institute of Computing currently manages only 16 spaces, the approach can be extended to more complex scenarios, such as the Unicamp rectory parking lot, which operates with a four-camera system and manages over 100 parking spots. To work with larger parking lots, the YOLOv11m (TFlite) model used in our parking lot may not provide the desired accuracy, and new options can be considered such as larger models or others such as YOLOv12 \cite{tian2025yolov12},  Transformers based real time models \cite{huang2024deim,peng2024d}, and vision foundation models \cite{xiao2024florence}.

\highlightyellowrevnew{The current analysis relies on some simple assumptions that contain limitations. The computational complexity of the spot-wise approach grows proportionally to the number of detections ($n$) and parking spots ($k$), resulting in an $O(n \times k)$ complexity for detection and analysis. This results in a quadratic complexity in the worst case scenario where the parking lot is full. Processing large parking lots could be computationally intensive, motivating the segmentation of large parking areas into smaller parking groups sectors to reduce the effective number of spots per computation. 
Another limitation is that the outlier detection based on $Z-scores$ assumes that the distribution of detected vehicle areas follows an Gaussian pattern, which might not hold true for all parking spots.}

\highlightyellowrevnew{The proposed system can be further enhanced by using more robust methods to identify detection errors at challenging spots based on the dimension areas of each detection. In addition, modeling the occupancy as a temporal sequence for each parking spot could allow the extraction of transition patterns between occupied and free states. Such temporal modeling could be combined with unsupervised learning models also based on the probability of classification and trained to discover challenging spots that frequently exhibit irregular detections, allowing targeted optimization of the detection model and more efficient allocation of monitoring resources.}

\hlsecondround{Regarding to the evolution towards a full} Digital Twin, future works include hosting the visualization at a webpage to make it available for the university community, and including more parking lots of the campus in this representation. Also, we aim to extend IC2 Parking with \hlsecondround{simulations and agent-based reasoning} capabilities, advancing it into a full digital twin. Monte Carlo methods will be employed to establish baseline occupancy patterns and to quantify system behavior under uncertainty, thereby supporting the evaluation of what-if scenarios such as capacity adjustments, demand fluctuations, policy interventions, and event-driven surges.
On top of this simulation layer, we envision the integration of an AI agent that acts as the orchestrator. This agent would function as a flexible and adaptive layer, capable of translating natural language queries into simulation parameters, executing and comparing multiple scenarios, integrating contextual information, and explaining results in human terms. In addition, it could forecast occupancy, detect sensor failures through anomaly detection, and continuously refine and update the digital twin.
Furthermore, by leveraging the NGSI-LD model to define and interlink different entities, it becomes possible to build a knowledge graph that, combined with a large language model, would enable more complex queries, strengthen decision support, and extend the simulation framework from a single parking lot to the scale of the entire campus.

On the application support server side, it can be enhanced with a more sophisticated OTA firmware update of the totem, and a further discussion of its hardware capabilities to host different support services to the applications. One TV Box could also be used as the edge device that performs inference, however this modification would imply modifying the box that encloses the device and replacing the current camera with a USB camera module.

\section*{acknowledgements}
We acknowledge our initial findings published in \cite{da202510} as a foundational baseline. This extended study introduces two new components: a digital twin and an application support server. Apart from that, we included a spot-wise approach with two methods that enhance the capabilities of YOLOv11m model, advancing the scope and performance beyond our previous work.
The authors would like to thank everyone who historically contributed to this project.

\section*{funding}
This research was supported by CAPES (process 88887.189925/2025-00), by the Brazilian Ministry of Science, Technology and Innovations, with resources from Law num. 8,248, of October 23, 1991, within the scope of PPI-SOFTEX, coordinated by Softex and published Arquitetura Cognitiva (Phase 3), DOU 01245.003479/2024-10, and by FAPESP (process 2023/00811-0). 

\section*{contributions}
All authors contributed to interpreting the results and writing and reviewing the manuscript. G.P.C.P.D.L. conducted the AI experiments. A.M.A.N conducted the Digital Twin experiments. T.G.B conducted the Applications Support Server implementation. All authors conducted the deployment of the totem. J.F.B. supervised the project.


\section*{interests}
The authors declare that they have no competing interests.

\section*{materials}

The code used in the experiments, along with instructions to build the hardware, is publicly available on GitHub\footnote{\url{https://github.com/discovery-unicamp/smartparking_unicamp/}}. Downsampling was performed during data collection to minimize identifiable content. Nonetheless, the full dataset is stored with restricted access at the Institute of Computing, Unicamp, in line with a conservative approach to potential privacy issues. While license plates are not visible, any that do appear in this work are blurred as a precautionary privacy measure.


\bibliography{references}  






\section*{Appendix - Area Distribution and Outlier Analysis}
\label{sec:appendix}

As described in Section \ref{subsec:abbp}, to investigate the distribution of detected vehicle areas for each parking spot and identify potential detections errors, we generated boxplots per spot using the recorded areas in pixels from all images of the dataset. Detection results were expanded so that each row contained a single detection area and its corresponding spot identifier. Areas were then grouped by spot number to produce a list of areas for each spot.

\begin{figure}[!htbp]
    \centering
    \includegraphics[width=0.44\textwidth]{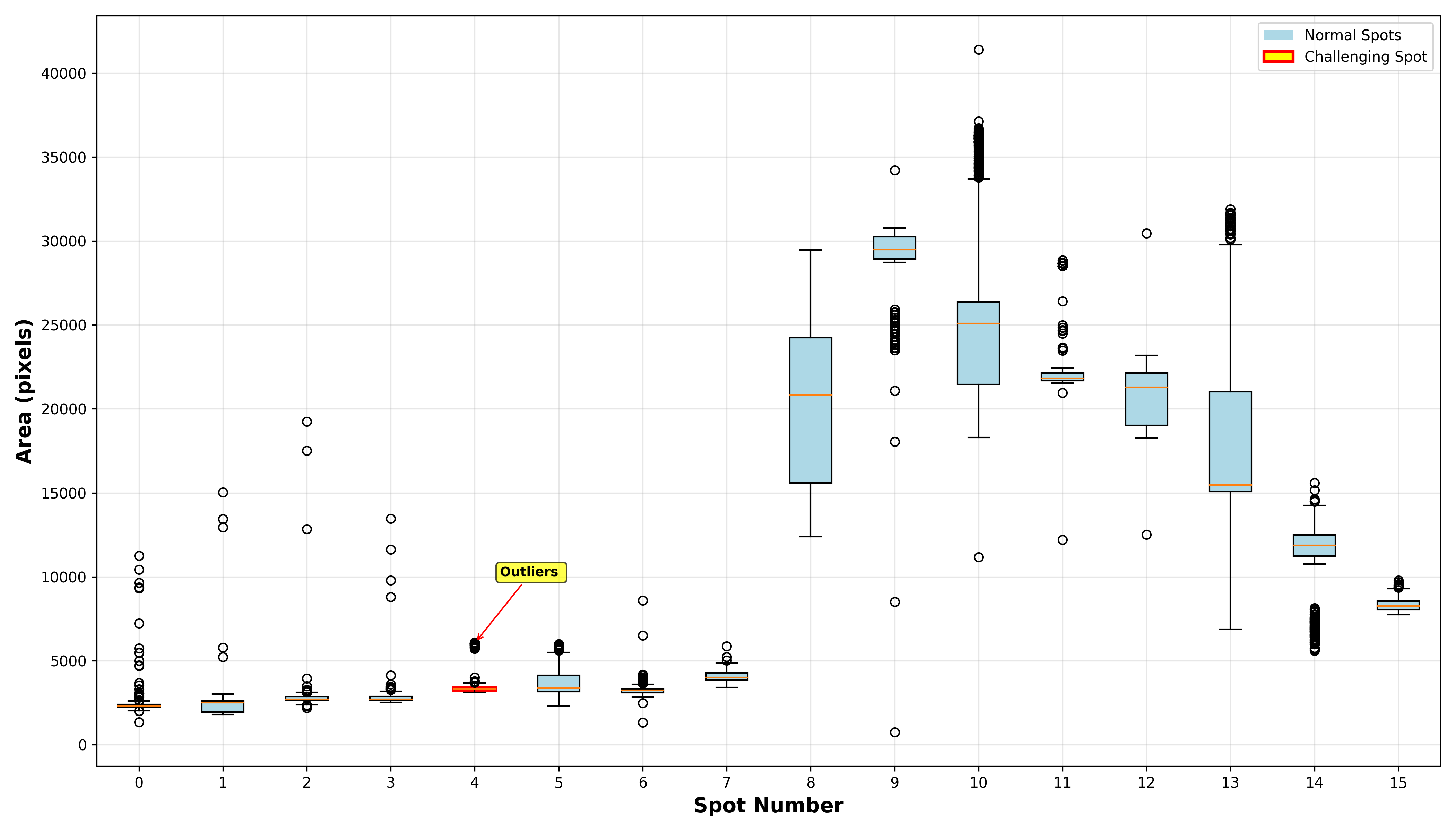}
    \caption{Boxplot summarizing the distribution of areas detected for each spot using the $Z-score$ method. Regular spots are shown in light blue, while the anomalous spot is highlighted in yellow with a red edge.}
    \label{fig:boxplot_spots}
\end{figure}

Figure~\ref{fig:boxplot_spots} shows a boxplot to summarize the distribution of detection areas for each spot , with regular spots shown in light blue and the spot identified with anomalies (spot 4) highlighted in yellow with a red edge. Outliers for spot 4 were identified using the $Z-score$ method, which measures how many standard deviations a given value is away from the mean. This approach is similar to the one previously applied to analyze variations in spot occupancy. In this case, the mean ($\mu$) and standard deviation ($\sigma$) of all detected areas for each spot were computed, and the normal range was defined as:

Figure~\ref{fig:boxplot_spots} shows a boxplot summarizing the distribution of detection areas for each spot, with regular spots shown in light blue and the anomalous spot highlighted in yellow with a red edge. Outliers for spot 4 were identified using the $Z-score$ method, which measures how many standard deviations a given value is away from the mean. This approach is similar to the one previously applied to analyze variations in spot occupancy, as it quantifies deviations relative to the local statistical behavior of each spot. In this case, the mean ($\mu$) and standard deviation ($\sigma$) of all detected areas for each spot were computed, and the normal range was defined as:
\[
\text{Lower bound} = \mu - 3\sigma, \qquad
\text{Upper bound} = \mu + 3\sigma.
\]
Any detection area falling outside this interval was flagged as an outlier and annotated on the boxplot. This visualization enables a straightforward comparison of detection area distributions across all spots and highlights regions with unusually large or inconsistent detections. Among all, spot 4 exhibited a wider dispersion and the highest amount of outliers above the upper limit, motivating a more detailed follow-up analysis.

\begin{figure}[!htbp]
    \centering
    \includegraphics[width=0.8\textwidth]{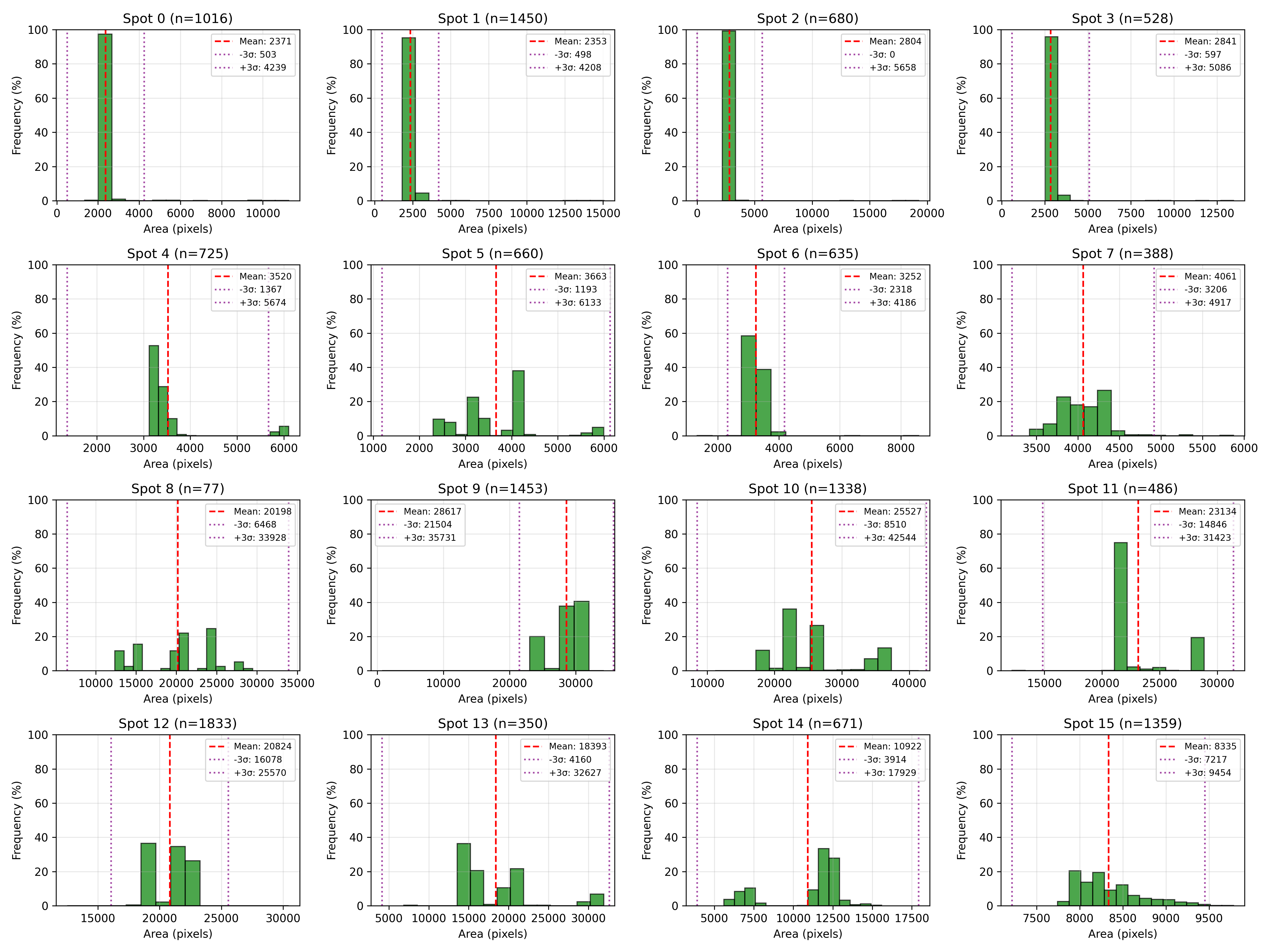}
    \caption{Spot-wise histograms of detected vehicle areas with mean,  and $\pm3\sigma$ highlighted.}
    \label{fig:spot_area_histograms_with_outliers}
\end{figure}

To further examine the detection area distributions and validate the results obtained from the boxplots, histograms were generated for all parking spots, as shown in Figure~\ref{fig:spot_area_histograms_with_outliers}. For each spot, we computed the mean and standard deviation of detected areas, plotting the corresponding $\pm3\sigma$ thresholds as vertical reference lines. When the lower bound was negative, it was adjusted to zero. This representation provides a clearer view of how detections are distributed around their mean and highlights the frequency and magnitude of deviations.

In this plot, we can observe that some spots like spot 5, 8, and 10 have areas distributions well confined within their expected range, while others such as spot 4 exhibit wider dispersion and clear outliers beyond the upper threshold. There are also cases where some outliers are present but the frequency of occurrence is small, such as spot 7. Interestingly, spot 11 showed a relatively higher number of outliers similar to spot 4, but when we visually inspected the corresponding images, these cases simply reflected normal variations in vehicle sizes rather than actual detection errors. This helped to select the $\pm3\sigma$ threshold to distinguish between genuine detection errors and natural variability among vehicles and parking spot sizes.

\end{document}